\documentclass[UTF8]{IEEEtran}
\usepackage{xcolor}
\usepackage{amsmath}
\usepackage[T1]{fontenc}% optional T1 font encoding
\usepackage{calligra}
\usepackage{algorithm}
\usepackage{algorithmicx}
\usepackage{algpseudocode}
\usepackage{amsmath}
\usepackage{caption}

\usepackage{cite}
\ifCLASSINFOpdf
   \usepackage[pdftex]{graphicx}
  \graphicspath{{../pdf/}{../jpeg/}}
  \DeclareGraphicsExtensions{.pdf,.jpeg,.png}
\else
  \usepackage[dvips]{graphicx}
  \graphicspath{{../eps/}}
  \DeclareGraphicsExtensions{.eps}
\fi
\usepackage{amsmath}
\usepackage[cmintegrals]{newtxmath}
\ifCLASSOPTIONcompsoc
  \usepackage[caption=false,font=normalsize,labelfont=sf,textfont=sf]{subfig}
\else
  \usepackage[caption=false,font=footnotesize]{subfig}
\fi
\usepackage{makecell}
\usepackage{amsmath}
\usepackage{soul}
\usepackage{bm}
\usepackage{multirow}
\usepackage{algorithm}
\usepackage{algpseudocode}
\usepackage{amsmath}
\usepackage{graphics}
\usepackage{epsfig}
\usepackage{booktabs}
\usepackage{diagbox}
\usepackage{setspace}

\begin{document}
%\onecolumn
%\title{semi-supervised Domain Adaptive Segmentation of Remote Sensing Images with Cross-Domain Multi-Prototypes Constraint and Contradictory Structure Learning}
%\title{Cross-Domain Multi-Prototypes with Contradictory Structure Learning for Semi-Supervised Domain Adaptive Segmentation of Remote Sensing Images}
%\title{\textcolor{red}{PCEL: A Prototype  and Context Enhanced Network for Domain Adaptation Semantic Segmentation of Remote Sensing Images}}
%\title{Class Asymmetry Domain Adaptation Learning for Remote Sensing Images Segmentation with Multiple Sources  }
\title{Integrating Multiple Sources Knowledge for Class Asymmetry Domain Adaptation Segmentation of Remote Sensing Images}
% note positions of commas and nonbreaking spaces  ( ~ ) LaTeX will not break

\author{~Kuiliang~Gao,~Anzhu~Yu,~Xiong~You,~Wenyue~Guo,~Ke~Li,~Ningbo~Huang% <-this % stops a space
\thanks{This work was supported in part by the National Natural Science Foundation of China under Grant 42130112, 42101458 and 41801388.}\thanks{~Kuiliang~Gao,~Anzhu~Yu,~Xiong~You,~Wenyue~Guo,~Ke~Li,~Ningbo~Huang, are with the PLA Strategic Support Force Information Engineering University, Zhengzhou,450001, China. (e-mail: youarexiong@163.com)
}% <-this % stops an unwanted space
}

% Spaces after \IEEEmembership other than the last one are OK  (and needed) as

\markboth{Journal of \LaTeX\ Class Files,~Vol.~14, No.~8, August~2015}%
{Shell \MakeLowercase{\textit{et al.}}: Bare Demo of IEEEtran.cls for IEEE Communications Society Journals}


%\IEEEspecialpapernotice{ (Invited Paper)}

\maketitle


% 遥感域适应问题中 存在着巨大域间差异 和 域内变化  导致了 复杂的类层次关系
% 如何更好的描述这些类层次关系 是关键
% 因此提出交叉域多原型方法
% 考虑到遥感影像同时存在大目标和小目标
% 设计了两种masked策略 增强语义信息学习 并且 为自监督训练提供高质量伪标签

\begin{abstract}
In the existing unsupervised domain adaptation (UDA) methods for remote sensing images (RSIs) semantic segmentation, class symmetry is an widely followed ideal assumption, where the source and target RSIs have exactly the same class space.
In practice, however, it is often very difficult to find a source RSI with exactly the same classes as the target RSI.
More commonly, there are multiple source RSIs available.
And there is always an intersection or inclusion relationship between the class spaces of each source-target pair, which can be referred to as class asymmetry.
Obviously, implementing the domain adaptation learning of target RSIs by utilizing multiple sources with asymmetric classes can better meet the practical requirements and has more application value.
To this end, a novel class asymmetry RSIs domain adaptation method with multiple sources is proposed in this paper, which consists of four key components.
Firstly, a multi-branch segmentation network is built to learn an expert for each source RSI.
Secondly, a novel collaborative learning method with the cross-domain mixing strategy is proposed, to supplement the class information for each source while achieving the domain adaptation of each source-target pair.
Thirdly, a pseudo-label generation strategy is proposed to effectively combine strengths of different experts, which can be flexibly applied to two cases where the source class union is equal to or includes the target class set.
Fourthly, a multiview-enhanced knowledge integration module is developed for the high-level knowledge routing and transfer from multiple domains to target predictions.
The experimental results of six different class settings on airborne and spaceborne RSIs show that, the proposed method can effectively perform the multi-source domain adaptation in the case of class asymmetry, and the obtained segmentation performance of target RSIs is significantly better than the existing relevant methods.

\end{abstract}


% \fi
\IEEEpeerreviewmaketitle

\section{Introduction}
\label{Sec_introduction}

\IEEEPARstart{R}{ecently}, the unsupervised domain adaptation (UDA) semantic segmentation of remote sensing images (RSIs) has attracted more and more attention \cite{9776640,9756442,rs14030646}.
Aiming to the knowledge transfer across RSIs domains, the UDA technologies can effectively improve the segmentation accuracy of unlabeled target RSIs, and greatly reduces the tedious workload of labeling RSIs.
Benefiting from the development of deep learning, various UDA methods have been proposed, constantly improving the performance of RSIs domain adaptation \cite{10105632,10113334}.
The existing methods can be roughly divided into three categories: RSIs style transfer, adversarial learning and self-supervised learning.
The style transfer methods take the lead in exploring RSIs domain adaptation, which mainly align source and target RSIs in the input space \cite{LUO2022105}.
Subsequently, with the continuous improvement of generative adversarial networks (GANs), it has become a popular solution to extract the domain-invariant features using adversarial learning \cite{9198144}.
More recently, the self-supervised learning methods have been gradually applied with the stable and efficient training process and superior performance, further improving the segmentation accuracy of target RSIs \cite{9745130}.

\begin{figure*}
\begin{center}
\resizebox*{0.85\linewidth}{!}{\includegraphics{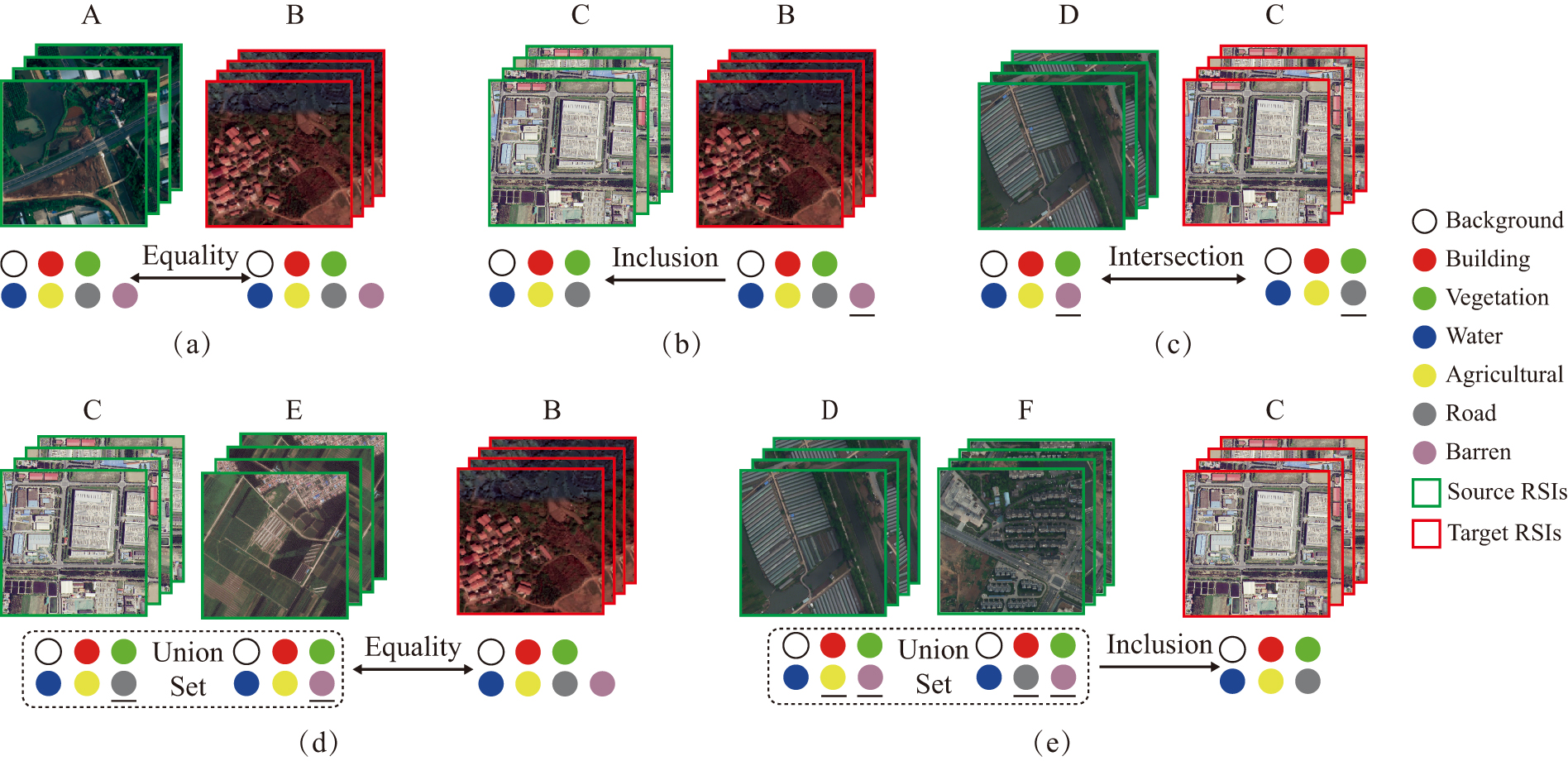}}\hspace{5pt}
\caption{The existing UDA methods of RSIs designed with (a) the class symmetry assumption cannot be effectively extended to the cases of (b) inclusion and (c) intersection between the source and target RSIs because of the asymmetry between class sets. Our work focuses on implementing class asymmetry domain adaptation segmentation of RSIs with multiple sources, in which the source class union (d) is equal to, or (e) includes the target class set.
}
\label{Fig_class_relationship}
\end{center}
\end{figure*}

Although there have been extensive researches on UDA semantic segmentation of RSIs, the existing methods generally follow an ideal assumption that labeled source domains and unlabeled target domains have exactly the same classes \cite{rs14174380}, which can be referred to as class symmetry.
As shown in Fig. \ref{Fig_class_relationship} (a), the source class set is equal to the target class set.
However, in practical application, it is often difficult and time-consuming to find a source RSI whose class set is completely consistent with that of target RSI.
More commonly, the class set of available source RSIs is different from that of target RSIs, and the two have inclusion or intersection relationship.
Fig. \ref{Fig_class_relationship} (b) shows the case where the target class set includes the source class set, and the latter is actually a subset of the former.
Fig. \ref{Fig_class_relationship} (c) illustrates the intersection relationship between the source class set and target class set, where each domain has a unique class that the other does not.
The above two relationships can be collectively referred to as class asymmetry, which is characterized by the fact that the target RSIs contain classes that do not appear in the source RSIs.
Considering the practical application situations of RSIs domain adaptation, the class asymmetry cases are obviously more common, but also more challenging.
However, the existing UDA methods following the class symmetry assumption cannot deal with the class asymmetry cases because of the inconsistency between source and target class sets.

No matter in the inclusion relationship of Fig. \ref{Fig_class_relationship} (b) or the intersection relationship of Fig. \ref{Fig_class_relationship} (c), the knowledge learned from source RSIs could not be well adapted and generalized to target RSIs because of the class asymmetry problem.
Indeed, the necessary condition for implementing RSIs domain adaptation is that the source class set includes the target class set.
In other words, any target class can be found in source domains, to ensure that the knowledge required by target RSIs can be learned from source RSIs.
Therefore, a natural idea for the class asymmetry domain adaptation is to collect more source RSIs, thus ensuring that the class union of multiple source domains is exactly equal to the target class set, or, under a laxer condition, includes the target class set.
Fig. \ref{Fig_class_relationship} (d) illustrates an example of the former, in which each source class set is a subset of the target class set, and the source class union has an equality relationship with the target class set.
Fig. \ref{Fig_class_relationship} (e) illustrates an example of the latter, in which each source class set intersects the target class set, and the source class union includes the target class set.
In both cases, the class spaces of different source RSIs are also different, which is close to the practical scenarios.
Undeniably, the introduction of more sources can provide the necessary information for multi-source RSIs domain adaptation in the case of class asymmetry.
However, in this novel and challenging experimental setup, there are still two key challenges that need to be addressed.

\textbf{Challenge 1:} In addition to distribution discrepancy, the class space between each source-target pair is also different, creating greater difficulties for knowledge transfer. Under this more complex condition, how to achieve the adaptation and alignment between each single source RSI and the target RSI is the first key challenge.

\textbf{Challenge 2:} There are distribution and class discrepancies among different source RSIs simultaneously.
Therefore, how to integrate the strengths and characteristics of multiple sources, complement each other, and efficiently transfer knowledge to target RSIs is the second key challenge.

Obviously, existing UDA methods of RSIs are struggling to address the above challenges, since they require a completely symmetric class relationship between source and target RSIs.
To this end, a novel Class Asymmetry Domain Adaptation method with Multiple Sources (MS-CADA) is proposed in this paper, which can not only integrate the diverse knowledge from multiple source RSIs to achieve better domain adaptation performance, but also greatly relax the strict restrictions of existing UDA methods.
For challenge 1, the proposed method utilizes a novel cross-domain mixing strategy to supplement class information for each source branch, and adapts each source domain to the target domain through collaborative learning between different sources RSIs.
For challenge 2, on the one hand, a multi-source pseudo-label generation strategy is proposed to provide self-supervised information for domain adaptation; on the other hand, a multiview-enhanced knowledge integration module based on the hypergraph convolutional network (HGCN) is developed for high-level relation learning between different domains, to fully fuse the different source and target knowledge and achieve better multi-source domain adaptation performance.
%Firstly, it provides a solution for class asymmetry RSIs domain adaptation, which has never been explored to our knowledge.
%Secondly, by integrating the rich knowledge of multiple sources, the better domain adaptation performance can be achieved and the segmentation accuracy of target RSIs can be further improved.
%Thirdly, the restrictive conditions of existing UDA methods are greatly relaxed, and the application scenarios and scope are effectively expanded.
To sum up, the main contributions include the following four points.
\begin{itemize}
\item[1)] A novel multi-source UDA method is proposed, which can effectively improve the performance of RSIs domain adaptation in the case of class asymmetry. To our knowledge, this is the first exploration of class asymmetry domain adaptation segmentation of RSIs.
\item[2)] A collaborative learning method based on the cross-domain mixing strategy is proposed, to achieve the domain adaptation between each source-target pair through supplementing class information for each source branch.
\item[3)] A pseudo-label generation strategy is proposed to deal with two different scenarios where the source class union is equal to or includes the target class set. A multiview-enhanced knowledge integration module is developed for efficient multi-domain knowledge routing and transfer by fully fusing the advantages of different branches.
\item[4)] Extensive experiments are conducted on airborne and spaceborne RSIs, and the results of three different scenarios, including two-source union equality, three-source union equality and two-source union inclusion, show that the proposed method can effectively perform the class asymmetry domain adaptation of RSIs with multiple sources, and significantly improves the segmentation accuracy of target RSIs.
\end{itemize}

\section{Related Work}

\subsection{RSIs domain adaptation segmentation}
\label{Sec_RSIs_UDA}

Accurately labeling RSIs is a very complicated and time-consuming work \cite{10017270,9769872,2020Gao}. To improve the generalization ability of deep segmentation models, more and more attention has been paid to the researches of RSIs domain adaptation.
From an intuitive visual perspective, the differences between RSIs mainly exist in color and other style attributes.
Therefore, the initial exploration mainly focuses on RSIs translation, aiming to reduce the discrepancies between the source and target domains by unifying the styles of different RSIs.
Tasar et al. present a color mapping GAN, which can generate fake images that are semantically exactly the same as real training RSIs \cite{9047180}.
Sokolov et al. focus on the semantic consistency and per-pixel quality \cite{9969990}, while Zhao et al. introduce the depth information to improve the quality of synthetic RSIs \cite{10004965}.
Only the implementation of style transfer in the input space often produces unstable domain adaptation performance.
Therefore, obtaining the domain-invariant deep features through adversarial learning has gradually become the mainstream in RSIs domain adaptation segmentation.
Cai et al. develop a novel multitask network based on the GAN structure, which possesses the better segmentation ability for low-resolution RSIs and small objects \cite{9999252}.
Zhu et al. embed an invariant feature memory module into the conventional adversarial learning framework, which can effectively store and memorize the domain-level context information in the training sample flow \cite{10038722}.
Zheng et al. improve the high-resolution network (HRNet) according to the RSIs characteristics, making it more suitable for RSIs domain adaptation.
In addition, the attention mechanism \cite{10032584,9667523}, contrastive learning \cite{9857935}, graph network \cite{9552486} and consistency and diversity metric \cite{10094018} are also integrated into the adversarial learning framework, further improving the segmentation accuracy of target RSIs.
Recently, the self-supervised learning based on the mean teacher framework \cite{WOS:000452649401023} has been gradually applied in RSIs domain adaptation due to its excellent knowledge transfer effect and stable training process.
Yan et al. design a cross teacher-student network, and improve the domain adaptation performance on target RSIs through the cross consistency constraint loss \cite{9385401}.
Wang et al. focus on the problem of spatial resolution inconsistency in self-supervised adaptation, effectively improving the effect of knowledge transfer from airborne to spaceborne RSIs \cite{WANG2022113058}.
In addition, combining the above different types of methods is a common idea to further improve the performance of RSIs domain adaptation \cite{10087276}.

Although various UDA methods for RSI semantic segmentation have sprung up, they all follow a common ideal assumption of class symmetry.
Under the condition that the source and the target class set are different, the existing methods often fail to achieve the satisfactory performance.

\subsection{RSIs multi-source domain adaptation}

The effective utilization of multiple RSIs sources can provide more abundant and diverse knowledge for improving RSIs domain adaptation performance.
However, most of the existing UDA methods of RSIs focus on the knowledge transfer from a single source to a single target domain, and there are relatively few researches tailored for multi-source RSIs domain adaptation.
The existing methods can be divided into two categories: constructing a combined source and aligning each source-target pair separately.
The former usually combines several different RSIs into a single source domain and performs cross-domain knowledge transfer according to the conventional UDA mode \cite{2021SSDAN,9137717,8935505,10.1117/1.JRS.15.036503}.
The obvious shortcoming is that the RSIs sources with different distribution in the combined domain will interfere with each other during domain adaptation process, thus weakening the effect of knowledge transfer \cite{WOS:000534424307032,WOS:000742075001020}.
In contrast, the latter typically aligns each source and target RSI separately, and merges the results from different sources as the final prediction of target RSIs \cite{rs10121890,9129743,WOS:000897090700001}.
The multiple feature spaces adaptation network proposed by Wang et al. is the closest to our work, which explores the performance of multi-source UDA in crop mapping by separately aligning each source-target RSI pair \cite{WOS:000897090700001}.
However, all the above work, including literature \cite{WOS:000897090700001}, require that each source and target RSI share exactly the same class space.
Obviously, meeting this requirement is even more laborious and cumbersome than the conventional UDA methods in Section \ref{Sec_RSIs_UDA}.

Our work focuses on the problem of RSIs class asymmetry domain adaptation. %and attempts to achieve ideal generalization on target RSIs by integrating multiple source knowledge.
Compared with existing multi-source UDA method, the proposed method can deal with the scenario closer to practical situation, that is, performing knowledge transfer using multiple RSIs sources with different class spaces.

\subsection{Incomplete and partial domain adaptation}

In the field of computer vision, there are two research directions related to our work, namely incomplete domain adaptation (IDA) and partial domain adaptation (PDA).
The problem setting of IDA is to utilize multiple sources with incomplete class for domain adaptation, and the existing few researches mainly focus on the image classification task \cite{WOS:000457843604012,WOS:000422952400006}.
Lu et al. and Gong et al. introduce IDA into the remote sensing field and preliminarily explore the performance of RSIs cross-domain scene classification \cite{8913683,9258408}, and Ngo et al. further deepen the research on this issue \cite{9828482}.
In addition, Li et al. propose to conduct the class-incomplete model adaptation without accessing source information, and design a deep model for street scene semantic segmentation \cite{WOS:000903735000033}.
However, the performance of this method on target domain is dissatisfactory, since it abandons the source knowledge in the domain adaptation process.
Different from the IDA setting, PDA is a single-source to single-target domain adaptation task.
However, in the PDA setting, the number of source classes is required to be greater than the number of target classes.
In the field of computer vision, various PDA methods have been developed for the image classification task \cite{WOS:000457843602088,9736609,INSPEC:22597845}.
Meanwhile, the performance of PDA methods on RSIs scene classification has begun to be studied, and the typical methods include the weight aware domain adversarial network proposed by Zheng et al. \cite{9984694}.

The differences between the existing IDA and PDA methods and our work can be summarized into the two aspects.
First, most of the existing methods are designed for the classification task of natural images, and the performance of these methods directly applied to the RSIs domain adaptation segmentation is greatly degraded.
Secondly, our work focuses on the class asymmetry domain adaptation of RSIs with multiple sources, and can simultaneously cover the two scenarios where the source class union is equal to or includes the target class set.
On the one hand, our work can directly perform RSIs domain adaptation with multiple class-incomplete sources; On the other hand, compared to the single-source PDA tasks, our work is more consistent with the practical situations where multiple sources with different class sets are available.

\section{Methodology}

\subsection{Problem setting}
\label{Sec_Problem_setting}

Formally, the problem setting of class asymmetry RSIs domain adaptation with multiple sources is first presented.
There are $k$ labeled source domains $\{\mathcal{D}_i^S\}^k_{i=1}$ with class sets $\{\mathcal{C}_i^S\}^k_{i=1}$ and one unlabeled target domain $\mathcal{D}^{T}$ with the class set $\mathcal{C}^{T}$.
The distribution $\mathcal{P}$ and class space $\mathcal{C}$ of each domain are known and distinct from each other, i.e., $\mathcal{P}^{S}_1\neq\mathcal{P}^{S}_2\ldots\neq\mathcal{P}^{S}_k\neq\mathcal{P}^{T}$, and $\mathcal{C}^{S}_1\neq\mathcal{C}^{S}_2\ldots\neq\mathcal{C}^{S}_k\neq\mathcal{C}^{T}$.
In addition, the source class union is equal to or contains the target class set, i.e., $ \mathcal{C}^T \subseteq \mathcal{C}^S_1\bigcup\mathcal{C}^S_2\ldots\bigcup\mathcal{C}^S_k$, and each source class set has an intersection with the target class set, i.e., $ \mathcal{C}^S_k \cap \mathcal{C}^T \neq \emptyset $.
Given source labelled samples $(x^{S_i}, y^{S_i})$ and target unlabelled samples $x^{T}$, the class asymmetry domain adaptation aims to achieve knowledge transfer from multiple sources to target RSIs.

\subsection{Workflow}

\begin{figure*}
\begin{center}
\resizebox*{0.9\linewidth}{!}{\includegraphics{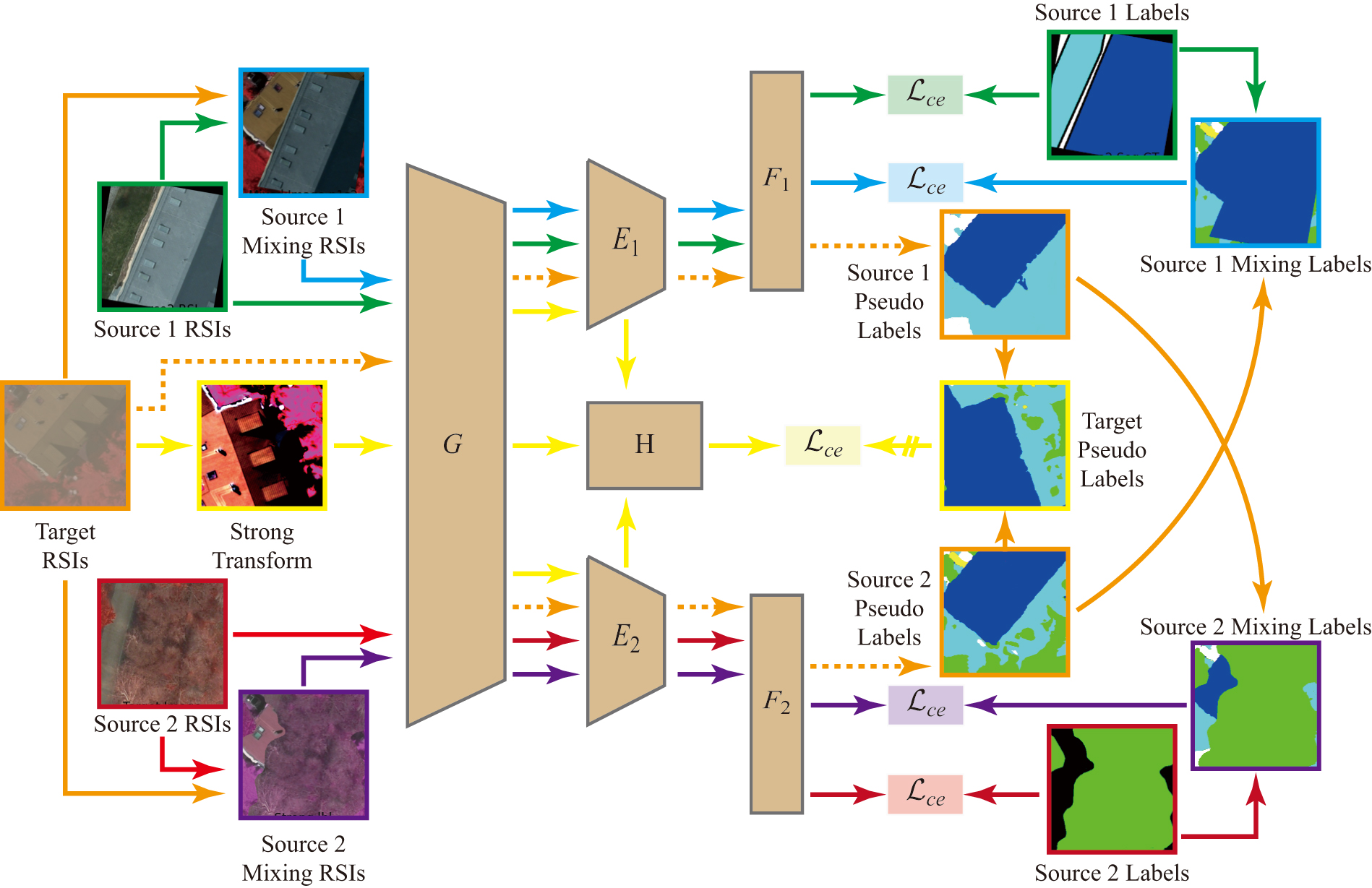}}\hspace{5pt}
\caption{Workflow of the proposed method. Different source RSIs are fed into their respective branches for supervised learning. For each source branch, the true labels are mixed with the target pseudo-labels of other branches, and the mixed results are used for collaborative learning and domain adaptation of each source-target pair. The final target pseudo-labels are generated from the predictions of different source branches, and the multi-domain knowledge is integrated for target inference. The dashed lines represent the data flows through the EMA model.
}
\label{Fig_workflow}
\end{center}
\end{figure*}

This paper proposes to integrate multiple sources knowledge for RSIs domain adaptation segmentation in the case of class asymmetry.
To clearly articulate the proposed MS-CADA method, the two-source scenario is used as an example to describe its workflow, as shown in Fig. \ref{Fig_workflow}.
The whole deep model $M$ consists of a feature extraction network $G$ and a multi-branch segmentation head.
The former is shared by multiple domains for common feature extraction existing in different RSIs, while the latter includes  two expert networks $E_1$ and $E_2$, two classifiers $F_1$ and $F_2$, and a knowledge integration module $H$.
The expert networks are used to learn the source-specific deep features, and the module $H$ is responsible for integrating knowledge from different domains and inferring target RSIs.
Overall, the proposed method follows the self-supervised adaptation mode, thus in each iteration, the teacher model $M'$ is established based on the exponential moving average (EMA) algorithm.
Specifically in each iteration, the proposed method performs three different learning tasks simultaneously, including supervised learning, collaborative learning, multi-domain knowledge transfer.

Firstly, the source RSIs and their corresponding true labels are used for model supervised training.
Due to the discrepancies in class space and data distribution, different RSIs are respectively fed into the corresponding expert networks and classifiers for loss calculation after passing through the shared $G$.
This step provides the basic supervision information for model optimization.
Secondly, to supplement the class information that each source branch does not have, the mixing of true labels and target pseudo-labels is performed between each source pair.
Correspondingly, the source and target RSIs are also mixed, and collaborative learning on multiple source branches is carried out according to the mixing results.
This step builds on the problem setting in Section \ref{Sec_Problem_setting}, where each source most likely contains class information that the other does not.
Thirdly, the predictions from different source branches in the teacher model are used to generate the final target pseudo-labels.
In the student model, the module $H$ fuses the deep features from different branches and performs multi-domain knowledge transfer based on the final target pseudo-labels.
In the following sections, the above learning process will be described in detail.

\subsection{Multi-source supervised learning}

Acquiring rich and robust source knowledge is a prerequisite for realizing class asymmetry multi-source domain adaptation.
Related researches have shown that simply combining different RSIs into one single source will lead to suboptimal domain adaptation effect due to the domain discrepancies \cite{8913683,9258408,9828482}.
Therefore, in each iteration, the RSIs from different sources will be fed into the corresponding expert branches, and supervised training will be conducted on the basis of learning common and source-specific features, which can be expressed as:
\begin{equation}
\label{equ1}
\mathcal{L}_{sup}^{S_i} = - \mathbb{E}_{(x_j^{S_i}, y_j^{S_i}) \sim D^S_i} \sum y_j^{S_i} log( F_i(E_i(G(x_j^{S_i}))) ),
\end{equation}
where $i$ and $j$ index different source domains and RSIs samples respectively in a training batch.
Utilizing different expert branches for supervised learning can effectively avoid the interference of domain discrepancies on model training, so as to provide support for the multi-source collaborative learning and knowledge integration in the next steps.

\subsection{Collaborative learning with cross-domain mixing}
\label{SEC_cross-domain_mixing}

Different source RSIs have different class sets.
Therefore, the multi-source supervised learning can only enable each expert branch to learn the knowledge within its corresponding class space.
In this case, the domain adaptation cannot be achieved due to the class asymmetry problem between each source-target pair.
To this end, a novel cross-domain mixing strategy is proposed to supplement each expert branch with the class information that it does not possess.

Specifically, the information supplementation from source 2 to source 1 is used as an example for detailed explanation.
As shown in Fig. \ref{Fig_workflow}, the class set of source 1 contains only Building, Low vegetation, and Impervious surface, thus during the initial phase of model training, the source 1 branch can only accurately recognize the three classes in target RSIs.
In contrast, the class set of source 2 contains the Tree class that source 1 does not have, and thus the target segmentation results of the source 2 branch will contain high-quality pseudo-labels of the Tree class.
Based on these observations, the proposed mixing strategy pastes part of the true label of source 1 onto the target pseudo-label generated by the source 2 branch, and the source 1 RSIs and the target RSIs are also mixed according to the same mask, which actually introduces the unseen class information into the source 1 branch.
Obviously, the mixing techniques play a key role in the proposed strategy.
In our work, the coarse region-level mixing \cite{WOS:000692171000136} and fine class-level mixing \cite{INSPEC:19399123} are adopted simultaneously.
Both the two techniques belong to the local replacement approaches, so they can be uniformly formalized as:
\begin{equation}
\label{equ2}
\begin{split}
x_{mix}^{S_1} = {\bf M} \odot x^{S_1}  + ({\bf 1}-{\bf M}) \odot x^T \\
y_{mix}^{S_1} = {\bf M} \odot y^{S_1}  + ({\bf 1}-{\bf M}) \odot y^T_{S_2},
\end{split}
\end{equation}
where $y^T_{S_2}$ denotes the target pseudo-labels of the source 2 branch of the EMA teacher model, the symbol $\odot$ represents the element-wise multiplication, and {\bf M} is a binary mask determining which region of pixels is cut and pasted.
When the class-level mixing is performed, the mask $\bf M$ is generated by randomly selecting partial classes in the true source labels, which delivers the inherent properties of a certain class of objects completely.
When the region-level mixing is performed, the mask $\bf M$ is obtained by randomly cutting a region patch from the true source labels, which can retain the local structure and context information.
Therefore, the application of the two mixing techniques can combine different advantages at the fine class and coarse region levels, effectively improving the performance of the information supplement between different sources.
In addition, the diversity of mixed samples is further enhanced, which can help to improve the robustness of the trained model.

After the mixed sample-label pairs $(x_{mix}^{S_1}, y_{mix}^{S_1})$ are obtained, the source 1 branch will carry out the weighted self-supervised learning.
Specifically, the confidence-based weight map is first generated as follows:
\begin{equation}
\label{equ3}
\begin{split}
&w^{S_1} = {\bf M} \odot {\bf 1}  +  ({\bf 1}-{\bf M}) \odot w_t \\
w_t = & \frac{ \sum_{l=1}^{h \cdot w} [ {\rm max}_c { F_1(E_1(G(x^{T}))) }^{(l,c)} > \tau ] }{h  \cdot w},
\end{split}
\end{equation}
where $h$ and $w$ denote the height and width of RSIs samples, the operation $[\cdot]$ is the Iverson bracket, and $w_t$ represents the pixel percentage in which the maximum softmax probability of prediction class exceeds the threshold $\tau$.
Then, the self-supervised loss of source 1 branch can be calculated as:
\begin{equation}
\label{equ4}
\mathcal{L}_{ssl}^{S_1} = - \mathbb{E}_{(x_{mix}^{S_1}, y_{mix}^{S_1})} \sum w^{S_1}   y_{mix}^{S_1} log( F_1(E_1(G( x_{mix}^{S_1} ))) ),
\end{equation}
which actually achieves the domain adaptation from source 1 to the target domain, since the input sample contains partial target RSIs.

\begin{figure}
\begin{center}
\resizebox*{0.9\linewidth}{!}{\includegraphics{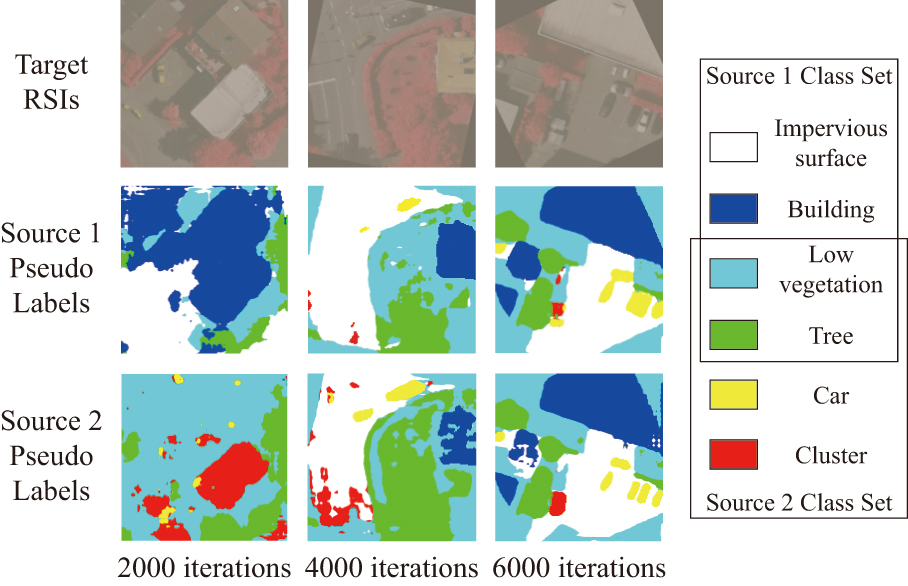}}\hspace{5pt}
\caption{Target pseudo-labels of different source experts.}
\label{Fig_mixing strategy}
\end{center}
\end{figure}

At this point, the process of class information supplementation from source 2 to source 1 has been completed.
Conversely, the information supplementation from source 1 to source 2 is similar to the above process.
To intuitively observe the effectiveness of the proposed cross-domain mixing strategy, the target pseudo-labels generated by different source experts are visualized, as shown in Fig. \ref{Fig_mixing strategy}.
Obviously, at the initial training stage, each source expert can only segment the target RSIs within its own class space.
For example, in the first column of Fig. \ref{Fig_mixing strategy}, the source 1 expert could not accurately identify the Car class, while the source 2 expert wrongly classifies the objects of impervious surface and building into the Cluster and Low vegetation classes.
With the continuous training, however, the class information among multiple source domains is supplemented with each other, and all the expert branches can identify all source classes more accurately.

When more source RSIs are adopted, the Equation \ref{equ2} will be extended to more forms including multiple mixed results of pairwise source combinations.
Summarily, in the proposed mixing strategy, each source can provide additional class information for other sources, which is actually the collaborative learning process between different experts.
As a result, each expert can learn all the class knowledge contained in the source union, thus effectively solving the class asymmetry problem during the domain adaptation process of each source-target pair.

\subsection{Knowledge integration for multi-domain transfer}

As stated in challenge 2, discrepancies between source RSIs domains can give different expert branches the strengthes of focusing on different feature knowledge.
After each source domain is adapted to the target domain separately, the advantages of different source experts should be combined to further improve the performance of multi-domain transfer.
Therefore, a multi-source pseudo-label generation strategy is proposed, which can deal with both cases where the source class union is equal to or includes the target class set.
And a multiview-enhanced knowledge integration module is developed for target inference through learning the high-level relations between different domains.

\subsubsection{Multi-source pseudo-label generation}

As shown in Fig. \ref{Fig_workflow}, in each iteration, the target RSIs will be fed into different expert branches of the EMA teacher model, which will output different predictions according to their respective ability of learning features.
Therefore, the final target pseudo-labels $y^T$ can be obtained:
\begin{equation}
\label{equ5}
\begin{split}
\hat{y}^T = {\rm max} ( F_1(E_1(G( x^{T} & ))), F_2(E_2(G( x^{T} ))) )\\
y^T =  M_c & ( \hat{y}^T ),
\end{split}
\end{equation}
where the $\rm max$ operation selects the class corresponding to the maximum softmax probability of the results of the two source branches.
And $M_c$ denotes the class filter operation, which is determined according to the relationship between the prediction class and target class set:
\begin{equation}
\label{equ6}
M_c = \begin{cases}
		 \hat{y}^T, \; \, {\rm if} \, \hat{y}^T \in \mathcal{C}^T ;\\
		 255, \, {\rm if} \, \hat{y}^T \notin \mathcal{C}^T.
\end{cases}
\end{equation}

\begin{figure}
\begin{center}
\resizebox*{0.7\linewidth}{!}{\includegraphics{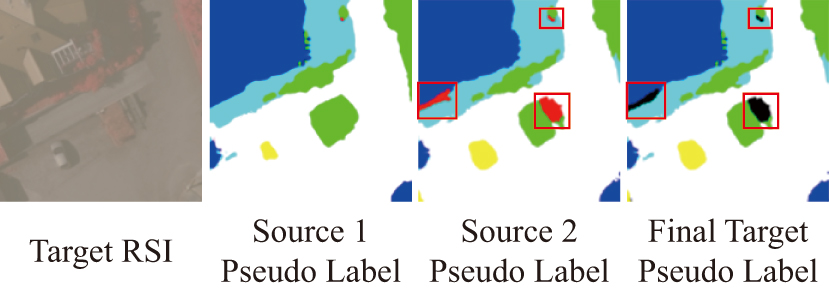}}\hspace{5pt}
\caption{Multi-source pseudo-label generation strategy. The pixels contained in the red box that correspond to the low-quality pseudo-labels at the boundary region are discarded.}
\label{Fig_pseudo_generation}
\end{center}
\end{figure}

In short, the proposed multi-source pseudo-label generation strategy has two effects.
(1) Equation \ref{equ5} can flexibly select the predictions of expert who is better at a certain class by comparing the confidence probabilities, which actually combines the advantages of different source branches.
(2) When the source class union contains the outlier classes that do not exist in the target RSIs, Equation 6 discards these classes by assigning them the value of 255 (pixels with the value of 255 are not used for loss calculation).
Actually, considering that the target RSIs do not contain the objects belonging to the outlier class, there will be a few target pixels classified into this class, which are often located in the boundary region between objects in target RSIs.
Therefore, the removal of these pixels from the final predictions can play a positive role in selectively preserving the high-quality pseudo-labels.
Fig. \ref{Fig_pseudo_generation} shows an example to visually explain the proposed strategy.

\subsubsection{Multiview-enhanced high-level relation learning}

\begin{figure}
\begin{center}
\resizebox*{1.0\linewidth}{!}{\includegraphics{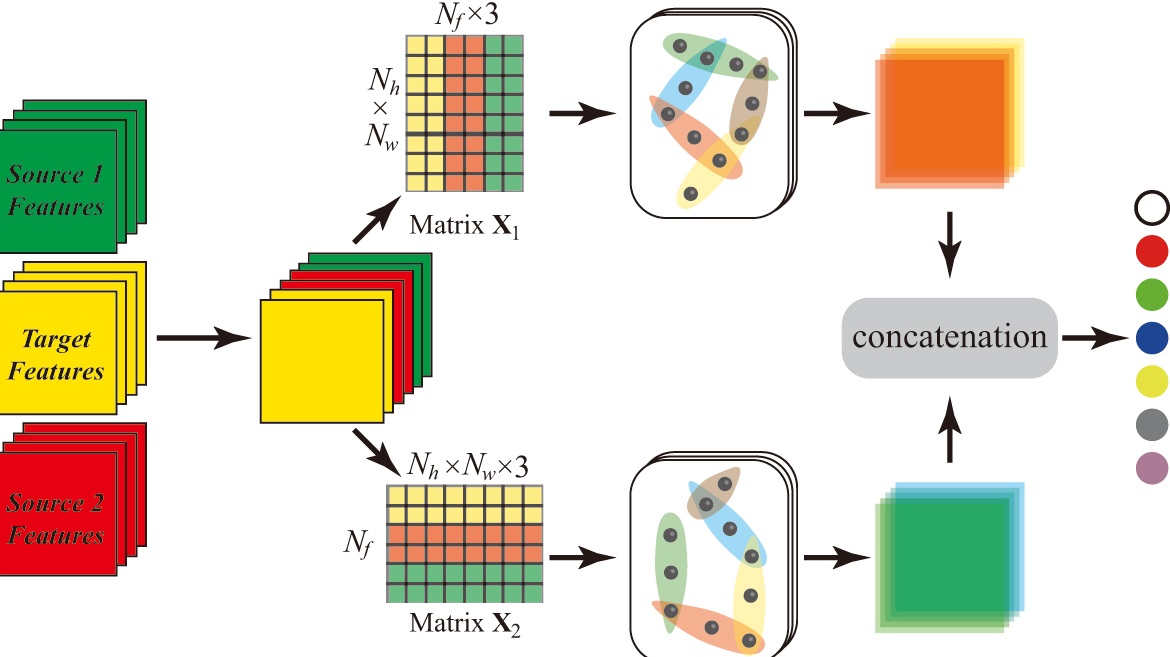}}\hspace{5pt}
\caption{Multiview-enhanced knowledge integration module based on hypergraph network for high-level relation learning.}
\label{Fig_hgcn}
\end{center}
\end{figure}

As shown in Fig. \ref{Fig_workflow}, for the input target RSIs after strong transformation, the features with source 1 bias learned by $E_1$, features with source 2 bias learned by $E_2$ and features learned by $G$ are fed into the module $H$ simultaneously, to realize the multi-domain knowledge transfer.
Theoretically, any network structure can act as the role of $H$.
Inspired by the recent development of HGCN in the computer vision field \cite{9878372,10049798,9590574}, a multiview-enhanced knowledge integration module based on HGCN is developed, as shown in Fig. \ref{Fig_hgcn}.
In the hypergraph structure $\mathcal{G}=(\mathcal{V}, \mathcal{E})$ where $\mathcal{V} = \{ v_1,...,v_n \}$ is the vertice set and $\mathcal{E} = \{ e_1,...,e_m \}$ is the hyperedge set, a hyperedge can connect arbitrary number of vertices simultaneously.
Therefore, superior to the abstraction of GCN on the pairwise connections or the modeling of CNN on the local features, the HCGN can better describe the high-level relations \cite{WOS:000585303400005,9795251}, which provides an effective solution for expressing the knowledge routing from different domains to the target predictions.

Firstly, the features learned by $G$ are compressed by a network with the same structure as $E_1$ and $E_2$, to ensure that the features from different branches have the same dimension $N_h \times N_w \times N_f$.
Then, the features of different branches are directly concatenated in the channel dimension.
Next, the high-level relation learning based on HGCN is carried out from two different views of space and feature.
The former refers to reshaping the obtained feature map into the $(N_h \times N_w) \times (N_f \times 3)$ matrix ${\bf X}_1$, where $N_h \times N_w$ corresponds to the number of pixels in the spatial dimension, and $N_f \times 3$ represents the richness of feature channels.
Consequently, the constructed hypergraph contains $N_h \times N_w$ vertices, each of which has the feature vector of $N_f \times 3$ dimension.
In this case, the hyperedges mainly describe the global contextual relation in the spatial dimension of feature maps.
And so on, based the matrix ${\bf X}_2$ of $N_f \times (N_h \times N_w \times 3)$, the latter will produce a hypergraph with $N_f$ nodes, each of which possessing the $N_h \times N_w \times 3$ feature vectors. The hyperedges model the high-level relation between different feature channels.
Finally, the outputs of hypergraphs with different views are concatenated and the target predictions are generated through a linear classification layer.
Obviously, implementing the high-level relation learning from the views of space and feature can more fully model the knowledge routing relation between different branches and target predictions, and effectively improve the effect of multi-source domain adaptation in the case of class asymmetry.

Next, the formal description of the above learning process and loss calculation is given.
For each vertex $v$ in the matrix ${\bf X}_1$ or ${\bf X}_2$, the $K$ nearest neighbor vertices are selected to build the hyperedge $e$, which can be denoted as:
\begin{equation}
\label{equ7}
e_i = \{ v_i, \forall v_j \in \mathcal{N}_K (v_i)  \},
\end{equation}
where $\mathcal{N}_K$ is actually the neighbor set containing $K$ vertices.
Specifically, the k-neighbor algorithm based on the Euclidean distance is used for generating the set $\mathcal{N}_K$.
After all possible hyperedges are built, the incidence matrix $\bf H$ can be obtained:
\begin{equation}
\label{equ8}
h(v,e) = \begin{cases}
		 1, \; {\rm if} \, v \in e ;\\
		 0, \; {\rm if} \, v \notin e,
\end{cases}
\end{equation}
where each entry describes the connection between vertices and hyperedges, and thus the matrix $\bf H$ can represent the whole topological structure of hypergraph.
Then, the hypergraph convolution operation is performed on the matrices ${\bf X}$ and ${\bf H}$:
\begin{equation}
\label{equ9}
{\bf Y} = \sigma (  {\bf D}_v^{-1/2} {\bf H} {\bf W} {\bf D}_e^{-1} {\bf H}^{\top} {\bf D}_v^{-1/2} {\bf X} {\bf \varTheta} ),
\end{equation}
where ${\bf Y}$ is the output results of hypergraph convolution, $\sigma$ is the ReLU activation function, ${\bf W}$ is the hyperedge weight matrix, and ${\bf \varTheta}$ is the trainable parameter matrix.
In addition, ${\bf D}_v$ and ${\bf D}_e$ denote the diagonal matrices of the vertex degrees and the edge degrees respectively, with each vertex degree calculated as $d(v) = \sum_{e \in \mathcal{E}} w(e) h(v, e)$ and each edge degree calculated as $\delta(e) = \sum_{v \in \mathcal{V}} h(v,e)$.

After the high-level learning by hypergraph convolution, the outputs of different hypergraph branches are reshaped into the feature maps with the spatial size of $N_h \times N_w$, and the channel concatenation and classification operations are performed to produce the prediction results $\bar{y}^T$ of target RSIs.
Therefore, the self-supervised loss for multi-domain knowledge transfer can be calculated as:
\begin{equation}
\label{equ10}
\mathcal{L}_{ssl}^{M} = - \mathbb{E}_{(\bar{y}^T, y^T)}  {\rm trans}(y^T) log( \bar{y}^T ),
\end{equation}
where the operation ${\rm trans}(\cdot)$ represents the same spatial transformation as in the strong transformation of input target RSIs.

\subsection{Optimization objective}

By combining the losses of multi-source supervised learning, collaborative learning and multi-domain knowledge transfer, the final optimization objective can be obtained:
\begin{equation}
\label{equ11}
\mathcal{L} = \mathcal{L}_{sup} + \alpha \mathcal{L}_{ssl} + \beta \mathcal{L}_{ssl}^M,
\end{equation}
where $\mathcal{L}_{sup}$ and $\mathcal{L}_{ssl}$ both contains the sum of the losses of different source branches, and $\alpha$ and $\beta$ denote the weight coefficients.
In addition, Algorithm 1 summarizes the pseudo code of the proposed MS-CADA method, to clearly show the entire workflow of class asymmetry RSIs domain adaptation in the two-source scenario.

\begin{algorithm}[htb]
\label{algorithm1}
\captionsetup{font=small}
\caption{MS-CADA for class asymmetry domain adaptation segmentation of RSIs with two sources}
\label{al_1}
\small
\begin{algorithmic}[1]
\Require $X^{S_1}, Y^{S_1}, X^{S_2}, Y^{S_2}$: labeled source RSIs samples
\Require $X^{T}$: unlabeled target RSIs samples
\Require $M$, $M'$:  student model and teacher model
\State randomly initialize $M$ and $M'$

\State \textbf{while} not done \textbf{do}
\State $\quad$ Sample batch ($x^{S_1}, y^{S_1}, x^{S_2}, y^{S_2}$) from ($X^{S_1}, Y^{S_1}, X^{S_2}, Y^{S_2}$)
\State $\quad$ Sample batch $x^{T}$ from $X^{T}$

\State $\quad$ \textbf{for} ($x^{S_1}, y^{S_1}, x^{S_2}, y^{S_2}, x^{T}$) \textbf{do}
\State $\quad$$\quad$ Calculate $\mathcal{L}_{sup}^{S_1}$ and $\mathcal{L}_{sup}^{S_2}$ with Equation \ref{equ1}

\State $\quad$$\quad$ Generate $x_{mix}^{S_1}, y_{mix}^{S_1}, x_{mix}^{S_2}, y_{mix}^{S_2}$ with Equation \ref{equ2}

\State $\quad$$\quad$ Calculate $\mathcal{L}_{ssl}^{S_1}$ and $\mathcal{L}_{ssl}^{S_2}$ with Equation \ref{equ4}

\State $\quad$$\quad$ Generate $y^{T}$ with Equation \ref{equ5}

\State $\quad$$\quad$ Calculate $\mathcal{L}_{sup}^{M}$ with Equation \ref{equ10}

\State $\quad$ \textbf{end for}

\State $\quad$ Calculate $\mathcal{L}$ with Equation \ref{equ11}

\State $\quad$ Update $M$ and $M'$

\State \textbf{end while}
\end{algorithmic}
\end{algorithm}

\section{Experimental Results}

\subsection{Datasets description and experimental setup}

\begin{table}[!t]
\footnotesize
\renewcommand{\arraystretch}{1.1}
\setlength\tabcolsep{4pt}
\caption{Details of different RSIs datasets.}
\label{table_RSIs}
\centering
\begin{tabular}{@{}ccccccccccc}
\toprule
Datasets& Subsets& Types & Coverage & Resolution & Bands\\%\diagbox{data set}{Architecture}
\midrule
ISPRS&VH&Airborne&1.38 $km^2$&0.09 m&IRRG\\
\midrule
ISPRS&PD&Airborne&3.42 $km^2$&0.05 m&\makecell{RGB\\IRRG\\RGBIR}\\
\midrule
LoveDA&\makecell{Rural\\Urban}&Spaceborne&536.15 $km^2$&0.3 m&RGB\\
\midrule
BLU&\makecell{Tile 1, Tile 2\\Tile 3, Tile 4}&Spaceborne&150 $km^2$&0.8 m&RGB\\
\bottomrule
\end{tabular}
\end{table}

The four public RSIs datasets, including ISPRS Potsdam (PD), ISPRS Vaihingen (VH), LoveDA \cite{DBLP:journals/corr/abs-2110-08733} and BLU \cite{9759447}, are used for experiments, and the details are listed in Table \ref{table_RSIs}.

\begin{table}[!t]
\footnotesize
\renewcommand{\arraystretch}{1.1}
\setlength\tabcolsep{3pt}
\caption{Different class settings in the two-source union equality scenario of airborne RSIs (PD1 + PD2 $\rightarrow$ VH).}
\label{table_Scenario1}
\centering
\begin{tabular}{@{}cccccccccccccccc}
\toprule
\multirow{2}{*} {\centering Class}  &  \multicolumn{3}{c}{Setting 1} &  \multicolumn{3}{c}{Setting 2} &  \multicolumn{3}{c}{Setting 3} \\
\cline{2-10}

&                             PD1& PD2 & VH                &        PD1& PD2 & VH               &        PD1& PD2 & VH  \\
\midrule
Impervious surface& $\checkmark$ &               &  $\checkmark$  &  $\checkmark$  &               & $\checkmark$  & $\checkmark$  & $\checkmark$  & $\checkmark$\\
Building&                        & $\checkmark$  &  $\checkmark$  &  $\checkmark$  &               & $\checkmark$  & $\checkmark$  & $\checkmark$  & $\checkmark$\\
Low vegetation&$\checkmark$      & $\checkmark$  &  $\checkmark$  &  $\checkmark$  & $\checkmark$  & $\checkmark$  & $\checkmark$  & $\checkmark$  & $\checkmark$\\
Tree&$\checkmark$                & $\checkmark$  &  $\checkmark$  &  $\checkmark$  & $\checkmark$  & $\checkmark$  & $\checkmark$  & $\checkmark$  & $\checkmark$\\
Car&$\checkmark$                 & $\checkmark$  &  $\checkmark$  &                & $\checkmark$  & $\checkmark$  & $\checkmark$  & $\checkmark$  & $\checkmark$\\
Clutter&$\checkmark$             & $\checkmark$  &  $\checkmark$  &                & $\checkmark$  & $\checkmark$  & $\checkmark$  & $\checkmark$  & $\checkmark$\\
\bottomrule
\end{tabular}
\end{table}

The VH and PD datasets contain the same six classes: Impervious surface, Building, Low vegetation, Tree, Car and Cluster.
Referring to relevant researches, the target VH dataset contains 398 samples for domain adaptation training and 344 samples for evaluation, while the source PD dataset contains 3456 samples for training \cite{9552486,10120939}.
The space size of each sample is $ 512 \times 512 $.
In the two-source union equality scenario of airborne RSIs, the whole PD training set is divided into two groups, i.e., the PD1 subset with 1728 RGB samples and the PD2 subset with 1728 IRRG samples.
As listed in Table. \ref{table_Scenario1}, the first two settings are obtained by discarding part of the classes in PD1 and PD2 respectively.
In addition, the class symmetry two-source setting is also used for experiments.

\begin{table}[!t]
\footnotesize
\renewcommand{\arraystretch}{1.1}
\setlength\tabcolsep{3pt}
\caption{Different class settings in the three-source union equality scenario of spaceborne RSIs (BLU1 + BLU2 + BLU3 $\rightarrow$ Urban).}
\label{table_Scenario2}
\centering
\begin{tabular}{@{}cccccccccccccccc}
\toprule
\multirow{2}{*} {\centering Class}  &  \multicolumn{4}{c}{Setting 1} &  \multicolumn{4}{c}{Setting 2} \\
%\midrule
\cline{2-9}
 &                             BLU1& BLU2& BLU3 & Urban       &       BLU1 & BLU2& BLU3 & Urban  \\
\midrule
Background& $\checkmark$    & $\checkmark$  &  $\checkmark$  &  $\checkmark$  & $\checkmark$  & $\checkmark$  & $\checkmark$  & $\checkmark$  \\
Building&$\checkmark$       & $\checkmark$  &  $\checkmark$  &  $\checkmark$  & $\checkmark$  & $\checkmark$  & $\checkmark$  & $\checkmark$  \\
Vegetation&$\checkmark$     & $\checkmark$  &  $\checkmark$  &  $\checkmark$  & $\checkmark$  & $\checkmark$  & $\checkmark$  & $\checkmark$  \\
Water&                      & $\checkmark$  &  $\checkmark$  &  $\checkmark$  & $\checkmark$  & $\checkmark$  & $\checkmark$  & $\checkmark$  \\
Agricultural& $\checkmark$  &               &  $\checkmark$  &  $\checkmark$  & $\checkmark$  & $\checkmark$  & $\checkmark$  & $\checkmark$  \\
Road&      $\checkmark$     & $\checkmark$  &                &  $\checkmark$  & $\checkmark$  & $\checkmark$  & $\checkmark$  & $\checkmark$  \\
\bottomrule
\end{tabular}
\end{table}

The class space merge is first performed on the LoveDA dataset by referring to \cite{9759447}, so that LoveDA and BLU datasets have the same six classes: Background, Building, Vegetation, Water, Agricultural and Road.
The Urban dataset used as the target domain contains 677 training samples and 1156 testing samples, and the first three tiles in the BLU dataset are used for different source domains, each of which contains 196 samples for domain adaptation training.
Each sample is cropped to the size of $1024 \times 1024$.
Similar to the scenario of airborne RSIs, the two three-source union equality scenarios are established, as listed in Table. \ref{table_Scenario2}.

\begin{table}[!t]
\footnotesize
\renewcommand{\arraystretch}{1.1}
\setlength\tabcolsep{3pt}
\caption{Different class settings in the two-source union inclusion scenario of airborne RSIs (PD1 + PD2 $\rightarrow$ Partial VH).}
\label{table_Scenario3}
\centering
\begin{tabular}{@{}cccccccccccccccc}
\toprule
\multirow{2}{*} {\centering Class}  &  \multicolumn{3}{c}{Setup 1} \\
\cline{2-4}
&                             PD1 & PD2 & Partial VH                  \\
\midrule
Impervious surface& $\checkmark$ &               &  $\checkmark$  \\
Building&$\checkmark$            &               &  $\checkmark$  \\
Low vegetation&$\checkmark$      & $\checkmark$  &  $\checkmark$  \\
Tree&$\checkmark$                & $\checkmark$  &  $\checkmark$  \\
Car&                             & $\checkmark$  &  $\checkmark$  \\
Clutter&                         & $\checkmark$  &                \\
\bottomrule
\end{tabular}
\end{table}

In addition, the two-source union inclusion scenario is established on the airborne RSIs, as shown in Table. \ref{table_Scenario3}.
Through discarding the RSIs containing the Clutter class of the VH dataset, the partial VH subset is obtained, where the number of training and testing samples is reduced to 350 and 319 respectively.

\subsection{Environment and hyperparameters}

All algorithms are developed based on Python 3.8 and relevant machine learning libraries.
A computer equipped with an Intel Xeon Gold 6152 CPU and an Nvidia A100 PCIE GPU provides hardware support for programs running.

Referring to most relevant researches, the ResNet-101 network pretrained on ImageNet dataset is used as the shared backbone $G$.
Each expert branch actually contains a improved atrous spatial pyramid pooling (ASPP) module as $E$ and a $1 \times 1$ convolution classification layer as $F$.
The number of output channels of $G$, $E$ and $F$ is 2048, 64 and $N_{\mathcal{C}^S}$ respectively, where $N_{\mathcal{C}^S}$ is the number of all source classes.
The HGCN in the knowledge integration module $H$ contains two hypergraph convolution layers.
In the view of space, the number of neighbor vertices used for building hypergraph is 64, and the number of channels of two layers is $64 \times k$ and 64 respectively, while in the view of feature, the above values are changed to 8, $N_h \times N_w \times k$ and $N_h \times N_w$ respectively.
In the multi-source union equality scenario, the liner classifier in $H$ outputs the $N_{\mathcal{C}^S}$ classes, while in the multi-source union inclusion scenario, it outputs the $N_{\mathcal{C}^T}$ classes.

During the domain adaptation training, the batchsize is set to 4, which means that there are $4 \times k$ source samples and $4$ target samples in a training batch.
For each source domain, half of the samples are mixed using the coarse region-level strategy, and the other half are mixed using the fine class-level strategy.
More specifically, in each source labeled sample, half of the classes or 40\% of the region is pasted to the target sample.
In the process of multi-domain transfer, the strong transformation of target RSIs containing flipping, rotation, cropping, color jitter and gaussian blur, and only the pixels with the prediction probability larger than 0.968 are used for loss calculation.
In addition, the Adam algorithm is used for model optimization, the training iteration is set to 40,000, and the learning rates of the backbone and multi-branch segmentation head are set to $6 \times 10^{-5}$ and $6 \times 10^{-4}$ respectively.
The weight coefficients $\alpha$ and $\beta$ are both set to 1.
Other related hyperparameter optimization techniques are consistent with \cite{9879466}.

\subsection{Methods and measures for comparison}

To verify the effectiveness of the proposed MS-CADA method, four conventional single-source UDA methods including Li's \cite{2022UDA_TRANS}, DAFormer \cite{9879466}, HRDA \cite{WOS:000903586400022} and PCEL \cite{10120939}, and four multi-source UDA methods including UMMA \cite{WOS:000903735000033}, DCTN \cite{WOS:000457843604012}, He's \cite{WOS:000742075001020} and MECKA \cite{9828482}, are used for performance comparison.

The UDA method presented by Li et al. introduces the two strategies of gradual class weights and local dynamic quality into the process of self-supervised learning, which can achieve better performance than most adversarial learning methods.
DAFormer is a simple and efficient UDA method, which can improve the target segmentation accuracy to a certain extent, by improving the training strategies of class sampling, feature utilization and learning rate.
HRDA is an improvement method over DAFormer, which can effectively combine the advantages of high-resolution fine segmentation and low-resolution long-range context learning.
PCEL is an advanced UDA method for RSIs segmentation, which can significantly improve the domain adaptation performance through the enhancement of prototype and context.
The above methods take the simple combination of different RSIs domains as the singe source, which can be referred to as "Combined source" for short.

Considering that there is no existing method that can exactly fit the problem setting of class asymmetry domain adaptation segmentation of RSIs, several multi-source UDA methods designed for related problems are modified appropriately and used for comparison.
UMMA is an advanced model adaptation method for street scene segmentation, which can achieve the adaptation to target domains using multiple models trained on different sources.
DCTN is an adversarial-based UDA method for natural images classification.
It can utilize different discriminators to generate the weights of different sources and combine different classifiers to achieve multi-source adaptation.
The method proposed by He et al. is actually a multi-source UDA method for street scene segmentation, and MECKA is a multi-source UDA method for RSIs scene classification based on the consistency learning between different domains.
Among them, only the methods UMMA and He's are capable of dealing with the multi-source union equality scenario of RSIs segmentation.
Therefore, appropriate modifications are imposed on the methods DCTN and MECKA.
Specifically, the classifiers of DCTN is replaced by the ASPP module, and the complete predictions are obtained through casting the results of different classifiers over the target class space and calculating the weighted results.
And the cross-domain mixing strategy proposed in Section \ref{SEC_cross-domain_mixing} is integrated into different branches of MECKA.
In addition, in the multi-source union inclusion scenario, for the above four methods, the outlier class that target RSIs do not contain will be discarded to re-form the multi-source union equality scenario, which is significantly different from the proposed MS-CADA method.
Consequently, the above methods can utilize multiple separated sources for class asymmetry UDA of RSIs, which can be referred to as "Separated multiple sources" for short.

All the above methods utilize the ResNet-101 trained on the ImageNet dataset as the backbone network, and perform domain adaptation training with the random seed of 0.
In addition, to fairly compare the performance of different methods, the three widely used measures including IoU per class, mIoU and mF1, are selected for quantitative comparisons.

\subsection{Results of the two-source union equality scenario of airborne RSIs}

\begin{table}[!t]
\footnotesize
\renewcommand{\arraystretch}{1.1}
\setlength\tabcolsep{2pt}
\caption{Segmentation results of the class setting 1 of the two-source union equality scenario. IoU per class is listed from column 3 to 8.}
\label{table_two_source_union_equality1}
\centering
\begin{tabular}{@{}ccccccccccccc}
\toprule
Type&Method&  \makecell{Imp.\\surf.} &  Build. & \makecell{Low\\veg.} & Tree & Car & Clu. & mIoU &mF1  \\%\diagbox{data set}{Architecture}
\midrule
\multirow{4}{*}{\makecell{Combined\\source}}&Li's&64.22&88.06&55.64&75.60&27.77&29.22&56.75&69.74\\
&DAFormer&67.75&82.15&50.40&61.03&60.00&15.73&56.18&69.33\\
&HRDA&72.39&83.78&50.26&66.46&50.23&44.48&61.27&73.08\\
&PCEL&71.12&87.62&54.89&66.62&47.07&45.78&62.18&73.91\\
\midrule
\multirow{4}{*}{\makecell{Separated\\multiple\\sources}}&UMMA&55.63&59.80&46.41&48.38&40.30&8.83&43.23&56.89\\
&DCTN&76.84&87.65&49.52&44.36&38.22&7.71&50.72&63.95\\
&He's&72.45&79.15&43.50&62.94&44.35&10.41&52.13&65.05\\
&MECKA&82.36&87.56&53.81&54.46&48.14&7.95&55.71&67.32&\\
&MS-CADA&77.39&89.47&56.45&67.34&51.94&47.06&64.94&76.26\\
\bottomrule
\end{tabular}
\end{table}

\begin{table}[!t]
\footnotesize
\renewcommand{\arraystretch}{1.1}
\setlength\tabcolsep{2pt}
\caption{Segmentation results of the class setting 2 of the two-source union equality scenario. IoU per class is listed from column 3 to 8.}
\label{table_two_source_union_equality2}
\centering
\begin{tabular}{@{}ccccccccccccc}
\toprule
Type&Method&  \makecell{Imp.\\surf.} &  Build. & \makecell{Low\\veg.} & Tree & Car & Clu. & mIoU &mF1  \\%\diagbox{data set}{Architecture}
\midrule
\multirow{4}{*}{\makecell{Combined\\source}}&Li's&78.98&88.39&62.26&74.57&9.83&24.00&56.34&69.13\\
&DAFormer&69.28&84.88&51.16&62.50&59.97&40.75&61.42&75.20\\
&HRDA&69.62&88.94&54.44&70.34&57.52&59.06&66.65&78.17\\
&PCEL&72.98&88.64&53.82&71.91&54.06&60.41&66.97&78.32\\
\midrule
\multirow{4}{*}{\makecell{Separated\\multiple\\sources}}&UMMA&52.73&56.17&40.59&55.20&37.30&10.23&42.04&55.91\\
&DCTN&79.42&87.54&50.28&44.96&38.57&7.66&51.41&64.54\\
&He's&78.07&85.17&52.81&53.07&48.00&17.58&55.78&67.88\\
&MECKA&81.54&86.27&53.28&56.20&52.00&15.88&57.53&69.96\\
&MS-CADA&80.25&89.68&55.02&66.24&55.59&75.20&70.33&81.39\\
\bottomrule
\end{tabular}
\end{table}

\begin{table}[!t]
\footnotesize
\renewcommand{\arraystretch}{1.1}
\setlength\tabcolsep{1.75pt}
\caption{Segmentation results of the class setting 3 of the two-source union equality scenario. IoU per class is listed from column 3 to 8.}
\label{table_two_source_union_equality3}
\centering
\begin{tabular}{@{}ccccccccccccc}
\toprule
Type&Method&  \makecell{Imp.\\surf.} &  Build. & \makecell{Low\\veg.} & Tree & Car & Clu. & mIoU &mF1  \\%\diagbox{data set}{Architecture}
\midrule
\multirow{5}{*}{\makecell{Single\\source}}&\makecell{DAFormer\\(PD1)}&64.13&87.36&27.63&30.48&64.53&1.52&45.94&57.14\\
&\makecell{DAFormer\\(PD2)}&82.56&89.81&55.83&65.16&64.45&47.34&67.53&79.71\\
&PCEL (PD1)&76.97&88.28&56.74&75.73&65.21&8.16&61.85&72.56\\
&PCEL (PD2)&81.51&89.55&59.40&72.32&57.96&59.45&70.03&81.07\\
\midrule
\multirow{4}{*}{\makecell{Combined\\source}}&Li's&81.95&88.75&55.77&65.38&58.99&40.34&65.20&77.83\\
&DAFormer&79.04&89.71&51.79&61.83&65.94&39.00&64.55&77.18\\
&HRDA&81.25&88.94&57.29&78.70&58.34&44.98&68.25&80.03\\
&PCEL&79.03&89.80&59.23&66.84&58.88&60.04&68.97&80.54\\
\midrule
\multirow{5}{*}{\makecell{Separated\\multiple\\sources}}&UMMA&68.37&70.19&47.26&56.71&45.63&16.38&50.76&64.82\\
&DCTN&79.31&88.12&53.85&65.29&62.85&15.05&60.75&73.26\\
&He's&79.58&86.16&55.79&62.48&56.46&36.28&62.79&76.05\\
&MECKA&79.51&88.65&54.72&67.81&64.75&50.26&67.62&79.13\\
&MS-CADA&83.73&90.08&60.15&68.40&59.89&70.93&72.20&83.36\\
\bottomrule
\end{tabular}
\end{table}

Tables \ref{table_two_source_union_equality1}-\ref{table_two_source_union_equality3} list the segmentation results of different methods in the two-source union equality scenario of airborne RSIs, which correspond to the three different class settings in Table \ref{table_Scenario1} respectively.
Observations and analysis can be obtained from the following four aspects.

Firstly, simply combining different RSIs and implementing domain adaptation based on the single-source UDA methods will weaken the segmentation performance of target RSIs.
According to the results of DAFormer and PCEL in Table \ref{table_two_source_union_equality3}, the mIoU based on the combined source decreases to different degrees, compared with the mIoU based on the best single source.
This is not difficult to understand that, different RSIs with domain discrepancies interfere with each other, resulting in the suboptimal domain adaptation performance.

Secondly, the performance of the four conventional single-source UDA methods with the combined source is obviously inferior to that of the proposed method.
Specifically, in the three different class settings, the mIoU of the proposed method is 2.76\%, 3.36\% and 3.23\% higher than that of the conventional UDA method with the best performance, respectively.
Correspondingly, the mF1 value of the proposed method is increased by 2.35\%, 3.07\% and 2.82\% respectively.
These observations directly verify the effectiveness of the knowledge integration and transfer with multiple separated sources, which effectively avoids the negative transfer problem in the combined source.

Thirdly, the results of four improved multi-source UDA methods and the proposed MS-CADA method are compared.
Specifically, UMMA can only utilize the models pretrained on different sources for domain adaptation without feature alignment between source and target RSIs, resulting in the relatively poor results.
DCTN can simply combine the advantages of different classifiers to some extent by weighted calculation, and its performance is better than UMMA.
The two methods He's and MECKA can perform the consistency learning and knowledge supplement between different source domains, and thus their domain adaptation performance is further improved.
The proposed method always performs better than the other four methods, wether in the class asymmetry case or the class symmetry case.
Compared to the second place, the proposed method improved by 9.23\%, 12.80\% and 4.58\% in mIoU and 8.94\%, 11.43\% and 4.23\% in mF1, in the three different class settings.
This demonstrates the superiority of the proposed methods in airborne RSIs domain adaptation segmentation in the two-source union equality scenario.

Fourthly, the performance of the proposed method in the class asymmetry scenario is still better than that of some advanced methods in the class symmetry scenario.
For example, the proposed method can achieve the higher mIoU and mF1 in class setting 2 than the other eight methods do in class setting 3, which fully demonstrates the effectiveness of the proposed method in integrating and transferring multi-source knowledge in the case of class asymmetry.

\begin{figure}
\begin{center}
\resizebox*{1.0\linewidth}{!}{\includegraphics{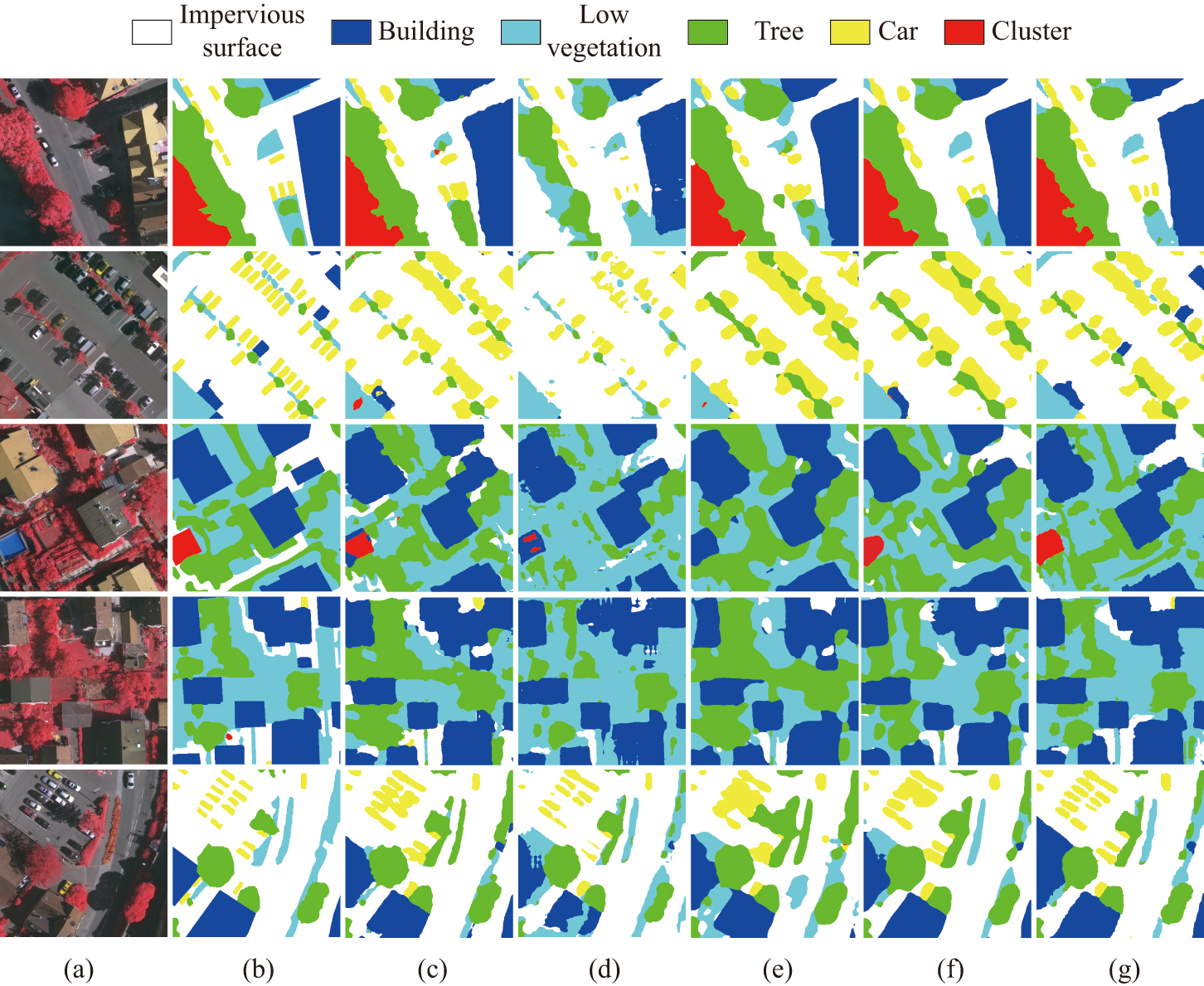}}\hspace{5pt}
\caption{Segmentation maps of different methods in the two-source union equality scenario of airborne RSIs. (a) RSIs. (b) Ground truth. (c) PCEL in class setting 3. (d) MECKA in class setting 2. (e) MECKA in class setting 3. (f) MS-CADA in class setting 2. (g) MS-CADA in class setting 3.}
\label{Fig_p2v}
\end{center}
\end{figure}

To intuitively compare the segmentation results of different methods, Fig. \ref{Fig_p2v} shows the segmentation maps of several examples in the target RSIs.
Specifically, the results of the three representative methods including PCEL, MECKA and MS-CADA in class settings 2 and 3 are visualized, from which the three main points can be drawn.
(1) The quality of the segmentation maps of the proposed method is superior to that of PCEL using the combined source.
Specifically, the segmentation maps of the proposed method possess the more complete context structure and less noise.
(2) Compared with MECKA, the proposed method can produce the more accurate segmentation maps, which is especially obvious for the minority class and small objects.
For example, in line 3, the proposed method can recognize the Clutter class accurately, and in line 5, the proposed method can locate the objects of the Car class more precisely.
(3) The segmentation maps in the setting 3 of class symmetry are superior to those in the setting 2 of class asymmetry, which can be clearly observed from Fig. \ref{Fig_p2v} (d)-(g).
Implementing RSIs domain adaptation with completely consistent class space can provide more class information and labeled samples for finely segmenting objects, so as to produce the segmentation maps with better quality.

\subsection{Results of the three-source union equality scenario of spaceborne RSIs}

% Background, Building, Vegetation, Water, Agricultural and Road

\begin{table}[!t]
\footnotesize
\renewcommand{\arraystretch}{1.1}
\setlength\tabcolsep{2pt}
\caption{Segmentation results of the class setting 1 of the three-source union equality scenario. IoU per class is listed from column 3 to 8.}
\label{table_three_source_union_equality1}
\centering
\begin{tabular}{@{}ccccccccccccc}
\toprule
Type&Method&  Back. &  Build. & Veg. & Water & Agri. & Road & mIoU &mF1  \\%\diagbox{data set}{Architecture}
\midrule
\multirow{4}{*}{\makecell{Combined\\source}}&Li's&27.37&47.97&18.79&10.69&21.12&42.79&28.12&42.75\\
&DAFormer&32.49&44.65&18.44&14.93&11.26&42.09&27.31&41.23\\
&HRDA&27.16&50.24&18.90&44.20&20.04&45.40&34.32&49.75\\
&PCEL&29.90&48.10&17.35&48.77&26.08&40.95&35.19&50.23\\
\midrule
\multirow{4}{*}{\makecell{Separated\\multiple\\sources}}&UMMA&19.55&30.98&13.13&13.76&2.86&20.51&16.80&29.37\\
&DCTN&25.96&43.54&16.95&19.31&14.30&35.20&25.88&39.04\\
&He's&24.30&47.75&12.31&46.67&16.60&34.26&30.32&44.69\\
&MECKA&29.83&50.21&18.82&49.89&15.19&44.61&34.76&49.86\\
&MS-CADA&30.82&52.76&23.48&43.29&22.36&46.70&36.57&51.69\\
\bottomrule
\end{tabular}
\end{table}

\begin{table}[!t]
\footnotesize
\renewcommand{\arraystretch}{1.1}
\setlength\tabcolsep{1.75pt}
\caption{Segmentation results of the class setting 2 of the three-source union equality scenario. IoU per class is listed from column 3 to 8.}
\label{table_three_source_union_equality2}
\centering
\begin{tabular}{@{}ccccccccccccc}
\toprule
Type&Method&  Back. &  Build. & Veg. & Water & Agri. & Road & mIoU &mF1  \\%\diagbox{data set}{Architecture}
\midrule
\multirow{5}{*}{\makecell{Single\\source}}&\makecell{PCEL\\(BLU1)}&30.83&47.76&20.68&56.25&29.11&45.12&38.22&53.37\\
&\makecell{PCEL\\(BLU2)}&27.03&37.81&18.72&62.74&26.64&44.55&36.25&51.04\\
&\makecell{PCEL\\(BLU3)}&31.49&47.27&19.21&54.32&30.06&48.08&38.41&53.63\\
\midrule
\multirow{4}{*}{\makecell{Combined\\source}}&Li's&40.83&46.77&17.50&30.95&23.88&51.33&35.21&50.94\\
&DAFormer&31.20&46.20&18.88&42.80&18.10&49.79&34.50&49.93\\
&HRDA&29.20&52.93&18.73&45.73&23.25&52.59&37.07&52.28\\
&PCEL&30.17&46.30&20.06&54.18&29.75&46.52&37.83&53.06\\
\midrule
\multirow{5}{*}{\makecell{Separated\\multiple\\sources}}&UMMA&18.70&32.22&13.71&27.60&5.04&31.89&21.53&35.65\\
&DCTN&26.52&44.68&18.38&27.91&18.66&42.22&29.73&34.92\\
&He's&21.03&48.71&17.48&42.38&14.02&49.32&32.16&37.47\\
&MECKA&23.40&50.40&18.88&54.68&21.55&50.04&36.49&51.59\\
&MS-CADA&31.82&48.34&20.95&58.04&26.23&54.26&39.94&55.36\\
\bottomrule
\end{tabular}
\end{table}

The segmentation results of different methods in the three-source union equality scenario of spaceborne RSIs are given in Tables. \ref{table_three_source_union_equality1}-\ref{table_three_source_union_equality2}.
Compared with airborne RSIs, the domain adaptation of large-scale spaceborne RSIs is more difficult, especially knowledge integration and transfer using multiple class asymmetry sources.
It can be seen that, there is a general decline in the segmentation performance of different methods, however, several observations similar to the airborne RSIs scenario can be obtained.

Firstly, according to the results of PCEL in Table \ref{table_three_source_union_equality2}, the performance based on the combined source is still inferior to that based on the best single source.
Although the expansion of training samples from three sources effectively reduces the decline range, the discrepancies between different RSIs sources still damage the segmentation performance in the target domain to some extent.
Secondly, in the two different class settings, the proposed method performs better than the best UDA method PCEL with the combined source, with the increase of 1.38\% and 2.11\% in mIoU, and 1.46\% and 2.30\% in mF1.
Thirdly, the segmentation results of the proposed method are better than those of other improved multi-source UDA methods.
Specifically, the proposed method improves mIoU by at least 1.81\% and 3.45\%, and mF1 by at least 1.83\% and 3.77\%.
It can be seen from the above statistics that, the proposed method can achieve better results than existing advanced single-source and multi-source UDA methods, whether in the class symmetry case or the class asymmetry case.

\begin{figure}
\begin{center}
\resizebox*{1.0\linewidth}{!}{\includegraphics{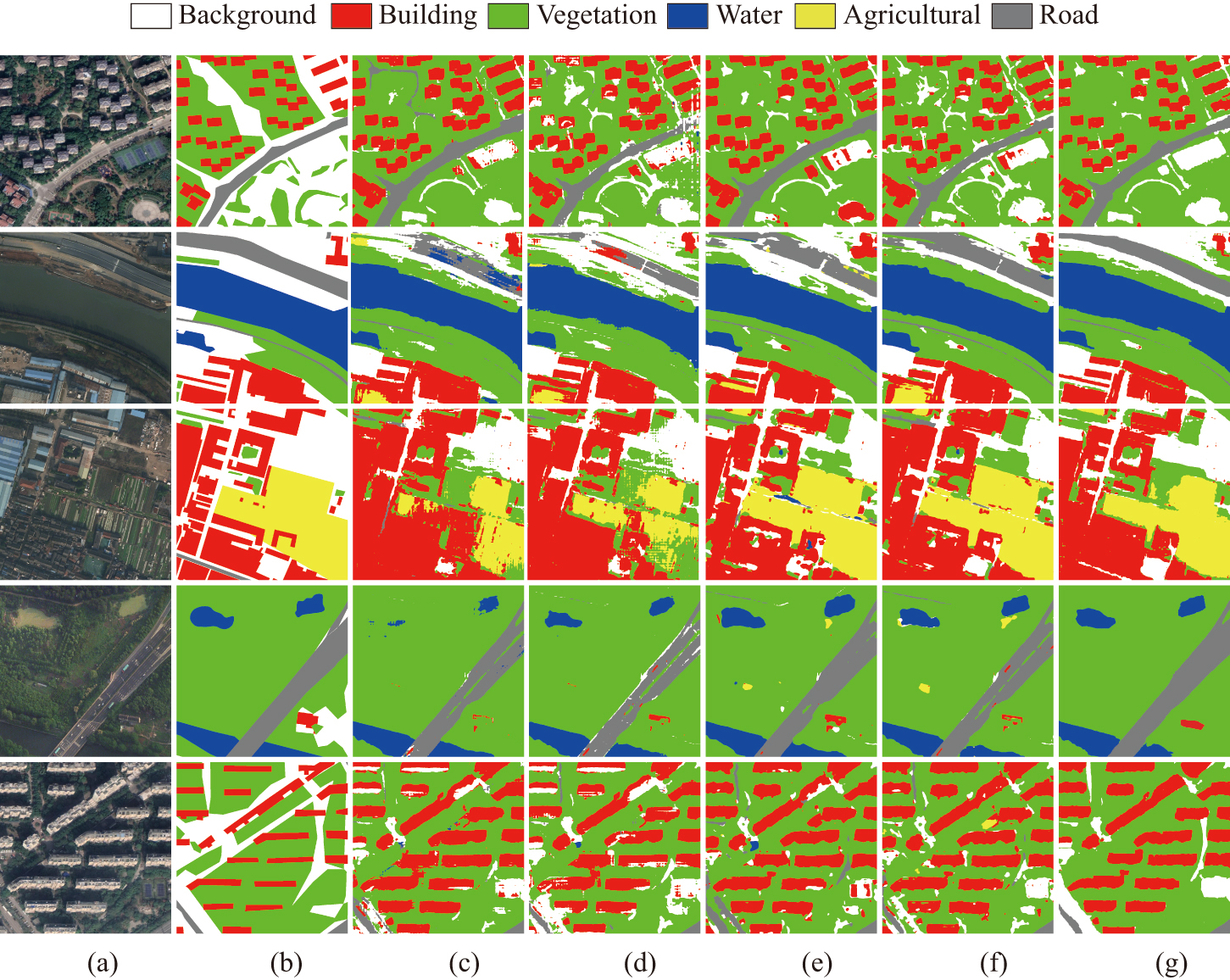}}\hspace{5pt}
\caption{Segmentation maps of different methods in the three-source union equality scenario of spaceborne RSIs. (a) RSIs. (b) Ground truth. (c) PCEL in class setting 2. (d) MECKA in class setting 1. (e) MECKA in class setting 2. (f) MS-CADA in class setting 1. (g) MS-CADA in class setting 2.}
\label{Fig_b2u}
\end{center}
\end{figure}

Fig. \ref{Fig_b2u} shows the segmentation maps of PCEL, MECKA and MS-CADA.
Compared with other methods, the segmentation maps of the proposed method in class setting 2 have the best visual effect and are closest to the ground truth.
Specifically, the recognition of the Building objects is more accurate and complete, and the edge of the segmentation results of linear objects is smoother.
In addition, under the same class setting, the segmentation maps of the proposed method are better than those of MECKA, which is mainly reflected in the less misclassification and noise.
This verifies the advantages of the proposed method in the three-source union equality scenario of spaceborne RSIs from the perspective of visualization.

\subsection{Results of the two-source union inclusion scenario of airborne RSIs}

\begin{table}[!t]
\footnotesize
\renewcommand{\arraystretch}{1.1}
\setlength\tabcolsep{2pt}
\caption{Segmentation results of the two-source union inclusion scenario. IoU per class is listed from column 3 to 7.}
\label{table_two_source_union_inclusion1}
\centering
\begin{tabular}{@{}ccccccccccccc}
\toprule
Type&Method&  \makecell{Imp.\\surf.} &  Build. & \makecell{Low\\veg.} & Tree & Car & mIoU &mF1  \\%\diagbox{data set}{Architecture}
\midrule
\multirow{4}{*}{\makecell{Combined\\source}}&Li's&68.75&87.57&54.71&63.54&19.38&58.79&67.78\\
&DAFormer&73.58&86.38&53.76&59.68&58.05&66.29&78.91\\
&HRDA&75.52&90.40&60.66&76.30&49.42&70.46&81.85\\
&PCEL&71.50&88.77&57.28&74.35&59.17&70.21&81.62\\
\midrule
\multirow{4}{*}{\makecell{Separated\\multiple\\sources}}&UMMA&63.26&68.12&27.02&56.91&22.92&47.65&60.13\\
&DCTN&67.61&80.73&47.75&58.36&42.71&59.43&70.55\\
&He's&79.97&85.91&50.03&44.04&57.94&63.58&75.84\\
&MECKA&81.91&88.27&53.40&52.05&62.45&67.62&79.76\\
&MS-CADA&82.25&90.90&57.54&65.67&62.85&71.84&83.37\\
\bottomrule
\end{tabular}
\end{table}

\begin{figure}
\begin{center}
\resizebox*{0.715\linewidth}{!}{\includegraphics{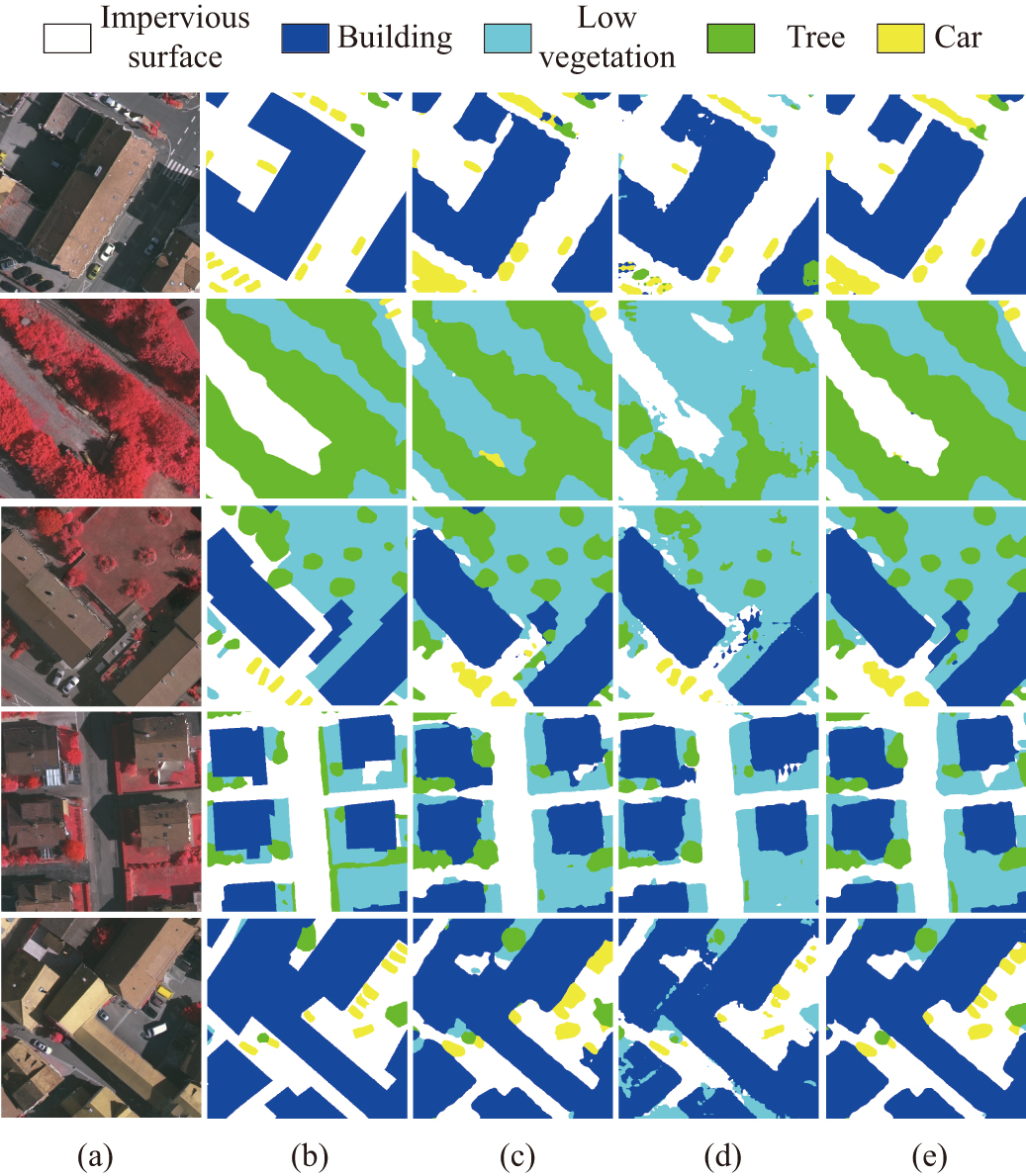}}\hspace{5pt}
\caption{Segmentation maps of different methods in the two-source union inclusion scenario of airborne RSIs. (a) RSIs. (b) Ground truth. (c) PCEL. (d) MECKA. (e) MS-CADA.}
\label{Fig_p2v_PDA}
\end{center}
\end{figure}

In addition to the multi-source union equality scenario, this section verifies the effectiveness of the proposed method in the two-source union inclusion scenario, and the results are listed in Table \ref{table_two_source_union_inclusion1}.
It should be pointed out that, for the eight UDA methods for comparison, the pixels corresponding to the Clutter class of the PD2 dataset in Table \ref{table_Scenario3} are discarded.
In other words, the eight UDA methods actually still perform domain adaptation in the two-source union equality scenario, because they cannot handle the source outlier classes that the target RSIs do not contain.

As listed in Table \ref{table_two_source_union_inclusion1}, the proposed method achieves the highest mIoU and mF1, which increase by 1.63\% and 1.75\% respectively compared with PCEL, and 4.22\% and 3.61\% respectively compared with MECKA.
It is believed that, such improvement benefits from the proposed multi-source pseudo-label generation strategy, which can effectively filter out the low-quality pseudo-labels located at the boundary between objects, and improves the performance of knowledge transfer in the self-supervised learning process.
In addition to the statistics, the segmentation maps of different methods are visualized, as shown in Fig. \ref{Fig_p2v_PDA}.
Compared with PCEL and MECKA, the proposed method can produce more fine segmentation maps.
For example, the segmentation results of small targets of the Car class and the edge of the Building objects are more accurate and detailed.

\section{Analysis and Discussion}

\subsection{Ablation studies}

\subsubsection{The cross-domain mixing strategy}

\begin{table}[!t]
\footnotesize
\renewcommand{\arraystretch}{1.1}
\setlength{\tabcolsep}{2pt}
\caption{Ablation studies of the proposed cross-domain mixing strategy (mIoU, \%).}
\label{table_ablation_1}
\centering
\begin{tabular}{@{}ccccccccccc}
\toprule
Scenario&\makecell{Class\\setting}& Baseline & \makecell{Class-level} &\makecell{Region-level} &Ours\\%\diagbox{data set}{Architecture}
\midrule
\multirow{3}{*}{\makecell{Two-source\\union equality}}&Setting 1&42.79&63.85&64.13&64.94\\
&Setting 2&41.55&69.80&68.94&70.33\\
&Setting 3&51.49&71.36&71.87&72.20\\
\midrule
\multirow{2}{*}{\makecell{Three-source\\union equality}}&Setting 1&21.38&35.51&35.79&36.57\\
&Setting 2&27.56&39.04&39.33&39.94\\
\midrule
\makecell{Two-source\\union inclusion}&Setting 1&49.87&71.06&70.95&71.84\\
\bottomrule
\end{tabular}
\end{table}

The proposed cross-domain mixing strategy can supplement the class information existing in other sources for each source branch, which is the key to achieve the collaborative learning among different branches and the adaptation from each source to the target RSIs.
The results of the ablation studies are presented in Table \ref{table_ablation_1}, where the Baseline method means that there is no mixing operation between different source branches.
As can be seen, the performance of the Baseline method is far behind that of other methods, since it can only implement multi-source domain adaptation using the pseudo-labels generated by the results with severe class bias.
In contrast, the performance brought by the class-level or region-level mixing strategy alone is significantly improved.
It is worth noting that, the region-level mixing can lead to higher mIoU in settings where class differences between sources are small, and the class-level mixing can lead to higher mIoU in settings where class differences between sources are large, such as the class setting 2 in the two-source union equality scenario.
The proposed strategy can absorb different advantages of class-level and region-level mixing, and simultaneously enhances the learning of fine inherent properties of objects and coarse local context structure.
Therefore, in the three different scenarios, the proposed strategy can enable the MS-CADA method to achieve the best performance.

\subsubsection{The multi-source pseudo-label generation strategy}

\begin{table}[!t]
\footnotesize
\renewcommand{\arraystretch}{1.1}
\setlength{\tabcolsep}{2pt}
\caption{Ablation studies of the proposed multi-source pseudo-label generation strategy (mIoU, \%).}
\label{table_ablation_2}
\centering
\begin{tabular}{@{}ccccccccccc}
\toprule
Scenario&\makecell{Class\\setting} & \makecell{Best expert} &\makecell{Summation} &\makecell{Ensemble} & Ours\\%\diagbox{data set}{Architecture}
\midrule
\multirow{3}{*}{\makecell{Two-source\\union equality}}&Setting 1&63.54&64.06&64.33&64.94\\
&Setting 2&68.87&69.73&69.61&70.33\\
&Setting 3&70.83&71.74&71.79&72.20\\
\midrule
\multirow{2}{*}{\makecell{Three-source\\union equality}}&Setting 1&35.02&35.63&35.86&36.57\\
&Setting 2&38.75&39.27&39.43&39.94\\
\bottomrule
\end{tabular}
\end{table}

How to generate the target pseudo-labels for multi-source adaptation based on the results of different expert branches is also a very key link in the proposed method.
The comparison of different pseudo-label generation strategies is shown in Table \ref{table_ablation_2}.
The Best expert method means that, only the results of the best expert branch are used for self-supervised training, without any integration operation on the different results of multiple experts.
In the Summation method, the logit results of different experts are firstly summed at the element level, and then the pseudo-labels are obtained through the softmax activation.
The Ensemble method comes from the research of Li et al. \cite{WOS:000903735000033}, and actually performs the selective average calculation on different results.
Through comparison, it is found that the self-supervised training with only the results of one source expert will produce the suboptimal performance.
Obviously, different experts will focus on different feature representation when performing supervised learning on different sources.
Therefore, integrating different strengths of source experts can effectively improve the performance of target tasks.
Compared with other methods, the proposed strategy can achieve the higher mIoU, which verifies its effectiveness in combining the advantages of different source experts in the process of multi-source domain adaptation.
In addition, it can deal with the scenario where the source class union includes the target class set, which other methods do not handle well.

\subsubsection{The multiview-enhanced knowledge integration module}

\begin{table}[!t]
\footnotesize
\renewcommand{\arraystretch}{1.1}
\setlength{\tabcolsep}{2pt}
\caption{Ablation studies of the designed multiview-enhanced knowledge integration module (mIoU, \%).}
\label{table_ablation_3}
\centering
\begin{tabular}{@{}ccccccccccc}
\toprule
Scenario&\makecell{Class\\setting}& CNN & Transformer & GCN &HGCN &Ours\\%\diagbox{data set}{Architecture}
\midrule
\multirow{3}{*}{\makecell{Two-source\\union equality}}&Setting 1&62.36&63.39&63.16&64.10&64.94\\
&Setting 2&68.14&68.67&68.72&69.26&70.33\\
&Setting 3&69.90&70.55&70.64&71.58&72.20\\
\midrule
\multirow{2}{*}{\makecell{Three-source\\union equality}}&Setting 1&35.18&35.82&35.07&35.75&36.57\\
&Setting 2&37.52&37.49&38.03&38.96&39.94\\
\midrule
\makecell{Two-source\\union inclusion}&Setting 1&69.87&70.47&70.32&71.05&71.84\\
\bottomrule
\end{tabular}
\end{table}

As described in the challenge 2 in the section \ref{Sec_introduction}, it is very important to fully fuse the feature information existing in different domains and realize efficient knowledge transfer in the process of class asymmetry RSIs domain adaptation.
This subsection explores the influence of different module $H$ on the domain adaptation performance, and the results are presented in Table \ref{table_ablation_3}.
The four typical deep models, CNN, Transformer, GCN and HGCN, are used for comparison.
They all have two layers, and the channel dimension of each layer is consistent with that of the designed module in the spatial view.
Except for the designed module, the other four models all perform feature learning based on the concatenation results of multi-domain features in the channel dimension.
Compared with CNN, Transformer and GCN, the HGCN model can obtain higher mIoU in target RSIs.
This validates the greater representation ability of the HGCN model, which is obtained by modeling the global many-to-many context relations.
The designed multiview-enhanced knowledge integration module can simultaneously perform the high-level relation learning in the views of space and feature, so it can realize better knowledge routing and transfer from different domains to target predictions.
According to the mIoU value, in the three different scenarios, the proposed multiview enhancement strategy brings at least 0.62\% improvement.

\subsection{Feature visualization}

%% examine the interpretability of deep models, such as feature visualization

\begin{figure*}
\begin{center}
\resizebox*{0.9\linewidth}{!}{\includegraphics{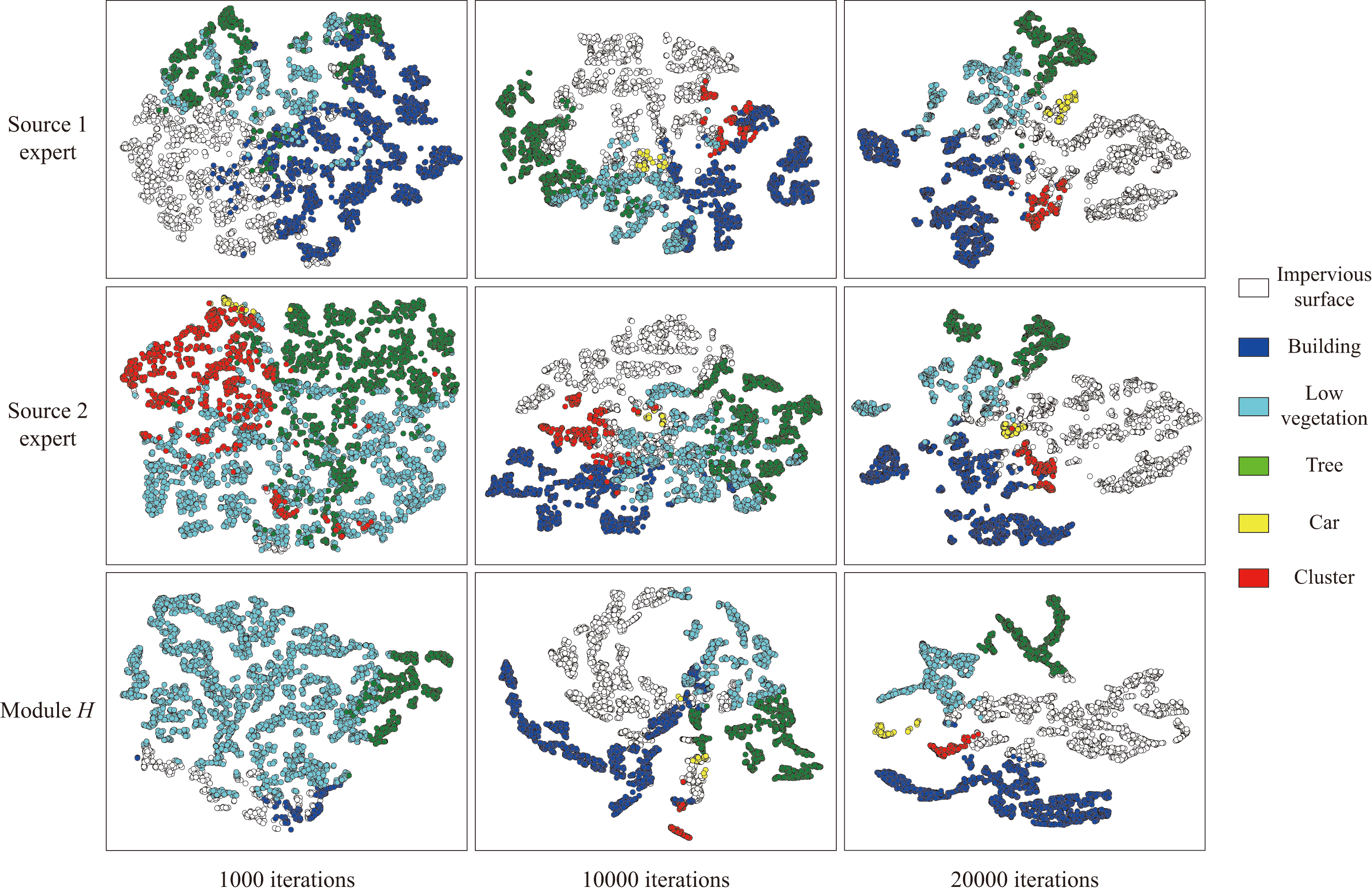}}\hspace{5pt}
\caption{Visualization of the deep features generated by different branches during RSIs domain adaptation.
}
\label{Fig_features}
\end{center}
\end{figure*}

To intuitively verify the effectiveness of the proposed method in the class asymmetry RSIs domain adaptation and better understand the process of class information supplement and multi-domain knowledge transfer, the visualization analysis is carried out for the deep features generated by different branches at different training iterations.
Fig. \ref{Fig_features} shows the distribution of features after the t-SNE dimension reduction \cite{article_tsne}.
Specifically, the same target RSI sample in the class setting 2 of the two-source union equality scenario is used as an example for analysis, and the features used for visualization are all derived from the stage before the classifiers in the source branches or in the module $H$.
Next, the detailed comparison and analysis are given from three different domain adaptation stages.

\subsubsection{1000 iterations}

As shown in the first column, in the initial training stage, each source branch maps the same target RSI into its own class space.
Therefore, the features produced by the source 1 expert are limited to the first four classes, while those produced by the source 2 expert are limited to the last four classes.
Such a mapping is bound to contain a lot of errors, and the ability of the module $H$ learned from these results is also relatively poor.
It can be seen that, there is a serious class bias problem in the feature distribution map of the module $H$.

\subsubsection{10000 iterations}

With the progress of domain adaptation training, the proposed cross-domain mixing strategy gradually comes into play.
As we can see, each source branch is supplemented with the initially unavailable class information, and thus it can more accurately map the target RSI sample into the full class space.
At the same time, the module $H$ can utilize the more accurate pseudo-labels for knowledge integration and transfer, and the class bias problem in the feature maps of different branches could be solved effectively.

\subsubsection{20000 iterations}

The whole model has been fully trained, and each source branch has a better ability to identify all classes.
Moreover, compared with the first two columns, the separability and discriminability of the features generated by source branches are significantly improved.
However, the mapping results of different source branches are still different, indicating that they have respective emphasis on different feature information.
Therefore, the model $H$ can draw complementary advantages from different domains, making the features belonging to the same class more clustered and the features belonging to different classes more separated.

From the variation of feature distribution with the process of domain adaptation training, it can be seen that the proposed cross-domain mixing strategy can complete the class information supplement, and achieve the domain adaptation of each source-target pair.
Moreoever, the designed module $H$ can effectively realize the knowledge transfer and class asymmetry multi-source adaptation based on the full utilization of different source features.

%% 在每个阶段对比源分支与模块H 后者更优

\subsection{Hyperparameters discussion}

\begin{figure}
\begin{center}
\resizebox*{1.0\linewidth}{!}{\includegraphics{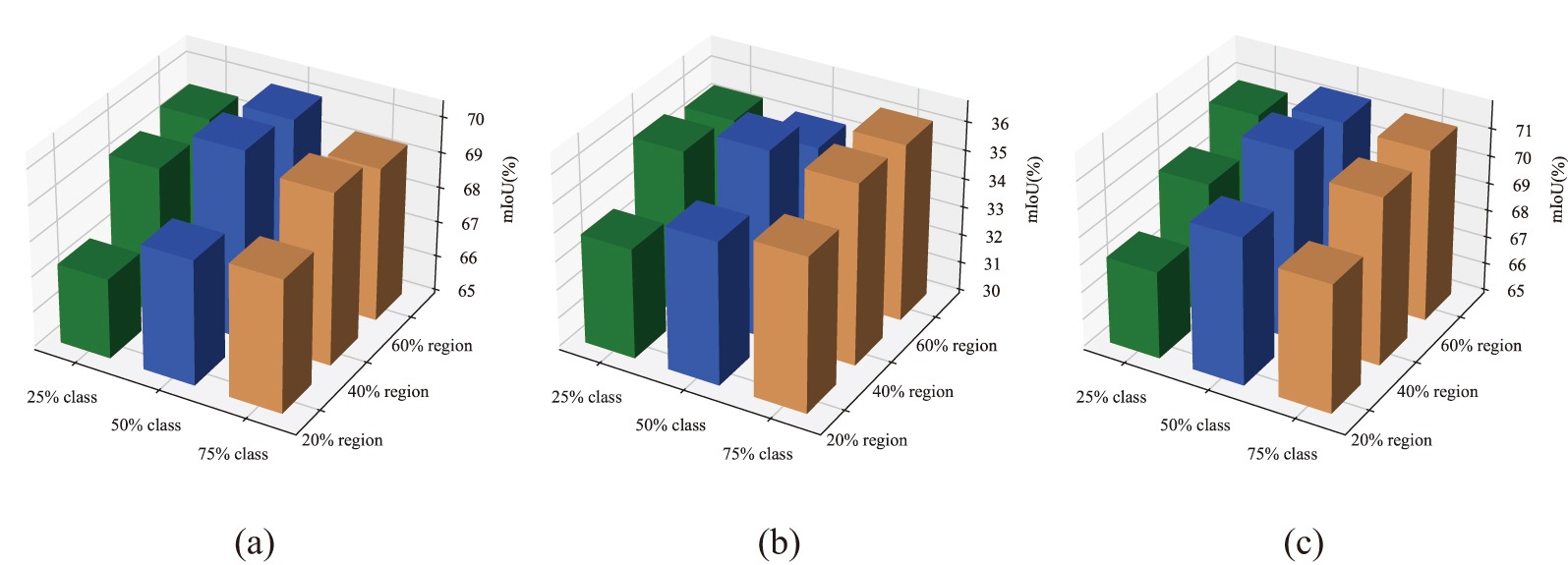}}\hspace{5pt}
\caption{Influence of different combinations of mixing ratios on segmentation results. (a) class setting 2 in the two-source union equality scenario. (b) class setting 1 in the three-source union equality scenario. (c) class setting 1 in the two-source union inclusion scenario.}
\label{Fig_3DBAR}
\end{center}
\end{figure}

In this section, the influence of three important hyperparameters on segmentation results is analyzed, to explore the sensitivity of the proposed method.
Firstly, the optimal combination of class mixing ratio and region mixing ratio in the proposed cross-domain mixing strategy is explored, and the results are shown in Fig. \ref{Fig_3DBAR}.
As we can see, when one ratio is too small, increasing the other ratio can lead to a significant improvement.
However, when one ratio is large enough, endlessly increasing the other ratio results in a slight decline in performance.
The purpose of constructing the mixed samples and labels is to supplement the feature information of other sources to a certain source branch and realize the domain adaptation of each source-target pair.
Therefore, the appropriate mixing ratios can better play the advantages of the proposed strategy.
Obviously, the best combination is "50\% class + 40\% region".

\begin{figure}
\begin{center}
\resizebox*{0.8\linewidth}{!}{\includegraphics{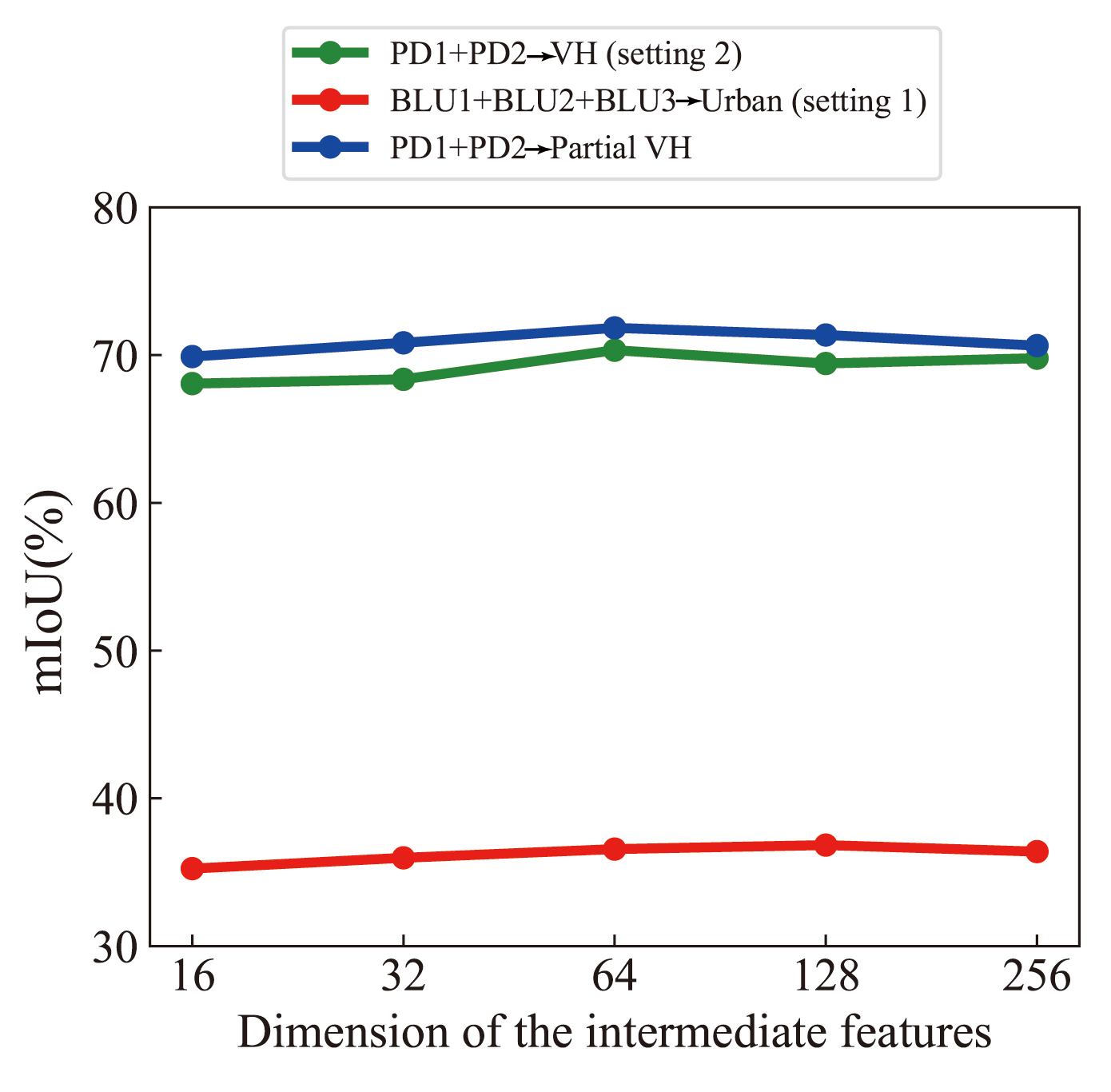}}\hspace{5pt}
\caption{Influence of the channel dimension of the intermediate features $N_f$ on segmentation results.}
\label{Fig_D}
\end{center}
\end{figure}

The channel dimension of the intermediate features $N_f$ determines the richness of information in the process of knowledge transfer from multiple domains to target predictions.
Specifically, the value of $N_f$ directly affects the structure and size of the constructed hypergraphes.
Fig. \ref{Fig_D} shows the influence of $N_f$ on the segmentation results of target RSIs.
On the whole, the curves corresponding to the three different scenarios show a trend of first rising and then stabilizing or even slightly declining.
Therefore, considering the performance and cost simultaneously, setting $N_f$ to 64 is a better choice for the experimental settings in this paper.

\begin{table}[!t]
\footnotesize
\renewcommand{\arraystretch}{1.1}
\setlength{\tabcolsep}{2pt}
\caption{Influence of different combinations of loss weights on segmentation results (mIoU, \%).}
\label{table_loss_weight}
\centering
\begin{tabular}{@{}ccccccccccc}
\toprule
Scenario& \makecell{Class\\setting} & \makecell{$\alpha = 1$,\\$\beta = 1$} &\makecell{$\alpha = 0.5$,\\$\beta = 1$} &\makecell{$\alpha = 1$,\\$\beta = 0.5$} & \makecell{$\alpha = 2$,\\$\beta = 1$} & \makecell{$\alpha = 1$,\\$\beta = 2$}  \\%\diagbox{data set}{Architecture}
\midrule
\makecell{Two-source\\union equality}&Setting 2&70.33&65.66&65.00&68.76&69.08\\
\midrule
\makecell{Three-source\\union equality}&Setting 1&36.57&34.19&33.96&35.73&36.04\\
\midrule
\makecell{Two-source\\union inclusion}&Setting 1&71.84&66.83&66.96&70.46&71.13\\
\bottomrule
\end{tabular}
\end{table}

Last but no least, the influence of the weight coefficients of different losses on the domain adaptation performance is analysed.
By observing the performance when $\alpha$ and $\beta$ are set to different values, the contribution of different components to the obtained results can be analyzed.
When one of $\alpha$ or $\beta$ is set to 0.5, the segmentation results of target RSIs are significantly worse, which indicates that the proposed collaborative learning method and the multi-domain knowledge integration module are very important for the proposed method to achieve excellent performance.
When one of A or B is set to 2, the performance is suboptimal, while setting both A and B to 1 can lead to the optimal performance.
This indicates that, it is better to treat $\mathcal{L}_{ssl}$ and $\mathcal{L}_{ssl}^M$ equally, because they respectively achieve the adaptation of each source-target pair and the knowledge aggregation of multiple domains, which together contribute to the significant improvement in the performance of class asymmetry RSIs domain adaptation with multiple sources.

\section{Conclusion}

Class symmetry is an ideal assumption that the existing UDA methods of RSIs generally follow, but it is actually difficult to be satisfied in practical situations.
Therefore, this paper proposes a novel class asymmetry RSIs domain adaptation method with multiple sources, to further improve the segmentation performance of target RSIs.
Firstly, a multi-branch segmentation network is built to conduct supervised learning of different source RSIs separately, which can effectively avoid the interference of domain discrepancies while learning the basic and rich source knowledge.
Then, the labels and RSIs samples are mixed simultaneously between different branches to supplement each source with the class information it does not originally have, and the collaborative learning among multiple branches is used to further promote the domain adaptation performance of each source to the target RSIs.
Next, the different advantages of source branches are combined for generating the final target pseudo-labels, which provides the self-supervised information for multi-source RSIs domain adaptation in the equality or inclusion scenario.
Finally, the knowledge aggregation module performs the multi-domain knowledge routing and transfer simultaneously from the views of feature and space, to achieve the better performance of multi-source RSIs domain adaptation.
The three scenarios and six class settings are established with the widely used airborne and spaceborne RSIs, where the experimental results show that the proposed method can achieve effective multi-source RSIs domain adaptation in the case of class asymmetry, and its segmentation performance in target RSIs is significantly better than the existing relevant methods.

Although the proposed method preliminarily explores the problem of RSIs domain adaptation with multiple class asymmetry sources, it is still limited to the transfer and adaptation of one target RSI domain.
In future work, the domain generalization, meta-learning and other related techniques will be introduced, to improve the adaptability of the proposed method in multiple target RSIs domains.

\ifCLASSOPTIONcaptionsoff
  \newpage
\fi



%\bibliographystyle{IEEEtran}
% argument is your BibTeX string definitions and bibliography database (s)
%  (used to reserve space for the reference number labels box)
\bibliography{mybibfile}   % bibliography data in report.bib
\bibliographystyle{ieeetr}   % makes bibtex use

\end{document}